\documentclass[10pt,twocolumn,letterpaper]{article}

\usepackage{cvpr}              % To produce the CAMERA-READY version
\usepackage[dvipsnames]{xcolor}
\usepackage{graphicx}
\usepackage{amsmath}
\usepackage{amssymb}
\usepackage{xfrac}
\usepackage{xcolor}
\usepackage[font=small,labelfont=bf,tableposition=top]{caption}
\usepackage{widetext}
\usepackage{capt-of}

\usepackage{multirow}
\usepackage{algorithm}
\usepackage{algpseudocode}
\usepackage{varwidth}
\usepackage[most]{tcolorbox}

\newcommand{\SX}{\mathcal{X}}  % Define a command for your special symbol

\newcommand{\SY}{\mathcal{Y}}
\newcommand{\RR}{\mathbb{R}}
\newcommand{\SL}{\mathcal{L}}
\definecolor{cvprblue}{rgb}{0.21,0.49,0.74}
\usepackage[pagebackref,breaklinks,colorlinks,citecolor=cvprblue]{hyperref}

\algrenewcommand\algorithmicrequire{\textbf{Input:}}
\algrenewcommand\algorithmicensure{\textbf{Output:}}

\def\paperID{10970} % *** Enter the Paper ID here
\def\confName{CVPR}
\def\confYear{2024}

\title{Pose-Guided Self-Training with Two-Stage Clustering for Unsupervised Landmark Discovery 
}

\author{Siddharth Tourani\textsuperscript{1,2}
\and
Ahmed Alwheibi\textsuperscript{1}
\and
Arif Mahmood\textsuperscript{3} 
\and 
Muhammad Haris Khan\textsuperscript{1} \qquad\qquad
\and
\small{
\textsuperscript{1}Mohamed bin Zayed University of Artificial Intelligence}, 
\textsuperscript{2}University of Heidelberg,
 \\
\small{\textsuperscript{3} Information Technology University of Punjab}\\
{\tt\small \href{tourani.siddharth@gmail.com}{tourani.siddharth@gmail.com}, \href{muhammad.haris@mbzuai.ac.ae}{muhammad.haris@mbzuai.ac.ae},
\href{arif.mahmood@itu.edu.pk}{arif.mahmood@itu.edu.pk}
}
}

\begin{document}
\maketitle
\begin{strip}
        \centering
\includegraphics[width=0.99\textwidth]{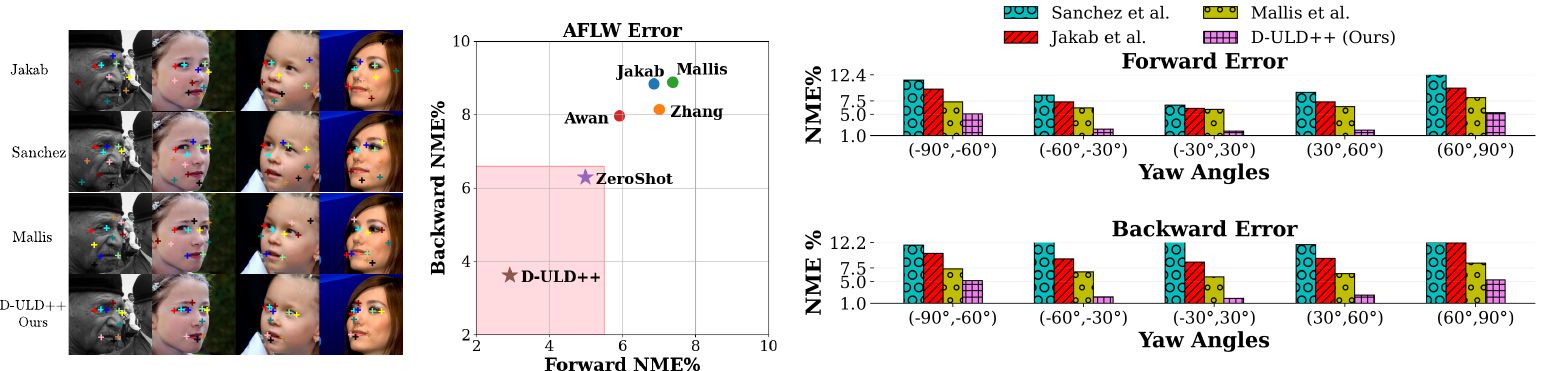}
\captionof{figure}{(Left) Visual comparison of proposed D-ULD++ with SOTA. 
(Middle) Mapping of various SOTA methods to NME space. (Right) D-ULD++ obtains minimum errors across yaw-angle ranges on AFLW Dataset. (Awan et al. \cite{awan2023unsupervised}, Jakab et al.\cite{jakab2018unsupervised}, Mallis et al.~\cite{mallis2023keypoints}, Sanchez et al.\cite{sanchez2019object},  Zhang et al. \cite{zhang2018unsupervised}).  The NME metrics are explained in~\cref{sec:experiments}.}
\label{fig:teaser}
\end{strip}

\vspace{-\baselineskip} % Adjust the space here

\begin{abstract}
\vspace{-0.4cm}
Unsupervised landmarks discovery (ULD) for an object category is a challenging computer vision problem.  
In pursuit of developing a robust ULD framework, we explore the potential of a recent paradigm of self-supervised learning algorithms, known as diffusion models. Some recent works have shown that these models implicitly contain important correspondence cues. Towards harnessing the potential of diffusion models for the ULD task, we make the following core contributions. 
First, we propose a ZeroShot ULD baseline based on simple clustering of random pixel locations with nearest neighbour matching. It delivers better results than  existing ULD methods. Second, motivated by the ZeroShot performance, we develop a ULD algorithm based on diffusion features using self-training and clustering which also outperforms prior methods by notable margins. Third, we introduce a new proxy task based on generating latent pose codes and also propose a two-stage clustering mechanism to facilitate effective pseudo-labeling, resulting in a significant performance improvement. Overall, our approach consistently outperforms state-of-the-art methods on four challenging benchmarks AFLW, MAFL, CatHeads and LS3D by significant margins. Code and models can be found at: \url{https://github.com/skt9/pose-proxy-uld/}. 

%we show that a simple clustering of internal representations of randomly sampled pixel locations with nearest neighbour matching delivers better results than the existing unsupervised landmark detection methods, (b) motivated by this, we develop an unsupervised landmark discovery algorithm based on diffusion features, self training and clustering which results in a strong baseline, which outperforms existing methods by notable margins, (c) we further capitalize on this strong baseline and introduce a new proxy task for unsupervised landmark detection via generating 3D pose as latent codes, (d) To effectively capture the variations while learning landmark correspondence, we leverage the 3D pose generated as latent code to devise a two-stage clustering mechanism, and (e) finally, our overall approach consistently outperforms state-of-the-art approaches on the challenging benchmarks of AFLW, MAFL, CatHeads and LS3D by over $50\%$ in forward and backward NME. 
\end{abstract}    

\section{Introduction}
\label{sec:intro}
\noindent\textbf{Background:} A large body of work approaches landmark detection for specific object categories in a fully-supervised setting  \cite{bulat2017far,khan2017synergy,wang2019adaptive,miao2018direct,dong2018style,kumar2020luvli}. They assume the availability of sufficiently annotated ground truth data. Common object categories for landmark detection include human faces or bodies as they offer an adequate quantity of images annotated with landmarks \cite{sanchez2019object}. However, akin to other computer vision problems, acquiring a large collection of annotated images for detecting landmarks in an arbitrary object category might be very costly \cite{mallis2023keypoints}. Therefore, in this paper, we aim to discover object landmarks in an unsupervised way.

\noindent\textbf{Challenges:} Unsupervised learning of object landmarks poses significant challenges due to several reasons.  Although landmarks are indexed with spatial coordinates, they convey high-level semantic information about object parts which is inherently difficult to learn without human supervision \cite{sanchez2019object, awan2023unsupervised}.   Detected landmarks should be invariant to different viewpoints, occlusions, and other appearance variations and also capture the shape perception of non-rigid objects, such as human faces \cite{sanchez2019object, mallis2023keypoints}. Some existing approaches to unsupervised landmark detection focus on learning strong representations, that can be mapped to manually annotated landmarks utilizing a few labeled images \cite{thewlis2017unsupervised_dense} or leverage proxy tasks, such as imposing equivariance constraints \cite{thewlis2017unsupervised, thewlis2017unsupervised_dense,thewlis2019unsupervised} or appending conditional image generation \cite{jakab2018unsupervised, sanchez2019object, zhang2018unsupervised, awan2023unsupervised}. 
Although obtaining promising performance in some cases, these methods struggle under different intra-class variations of an object, such as large changes in pose and expression (see Fig.~\ref{fig:teaser}).

\noindent\textbf{Motivation:} To deal with these challenges in unsupervised landmark discovery, we explore the potential of  pre-trained diffusion-based generative models \cite{ho2020denoising, song2020score}. Recent works have shown the utility of diffusion models beyond image synthesis, in tasks such as image editing \cite{meng2021sdedit, brooks2023instructpix2pix} and image-to-image translation \cite{parmar2023zero}. As such, these models are capable of converting one object into another object without modifying the pose and context of the former \cite{tang2023emergent}. This suggests the presence of implicit correspondence cues in the internal representations of diffusion models for different object classes \cite{luo2023diffusion,tang2023emergent, zhang2023tale}. %Such correspondence information

\noindent\textbf{Contributions:} 
\begin{itemize}[topsep=-3pt, noitemsep]
    \item To explore diffusion models for unsupervised landmark detection, we begin with a simple clustering of their pretrained internal representations indexed at randomly sampled pixel locations from object RoIs. For inference, we use nearest neighbour querying of randomly sampled pixel locations from unseen object image for discovering landmarks. Surprisingly, this \emph{zero-shot baseline} surpasses most existing methods.
%a  simple clustering of the zero-shot internal representations of the diffusion model, corresponding to randomly sampled pixel locations from object RoIs, followed by nearest neighbour querying of randomly sampled pixel locations of an unseen object image for discovering landmarks. 
 \item Motivated by the zero-shot baseline, we develop an \emph{unsupervised landmark detection algorithm built on diffusion features}, namely D-ULD, that uses clustering for pseudo-labelling which are then used for self-training. %to learn correspondences across object images. 
We note that, this simple algorithm produces superlative results and bypasses competing methods by visible margins. 

\item We capitalize on D-ULD and introduce \emph{a new pose-guided proxy task} which reconstructs landmark heatmaps after producing  latent pose codes. To better capture pose variations in landmark representations while clustering, we leverage these latent pose codes to develop \emph{a two-stage clustering mechanism} for better pseudo-labels, resulting in further performance improvement. The inclusion of pose-guided proxy task and two-stage clustering in D-ULD results in a robust ULD algorithm, namely D-ULD++ (see Fig.~\ref{fig:teaser}). 

\item Extensive experiments on four challenging datasets, featuring human  and cat faces, reveal that D-ULD++ consistently achieves remarkable performance across all datasets. 
\end{itemize}
%Moreover, an important feature of our method is the \emph{unsupervised generation of latent pose code} which can be exploited in various ways to induce auxiliary information into the unsupervised landmark detection pipeline.  

\begin{figure*}[!th]
    \centering
    \includegraphics[width=0.99\textwidth]{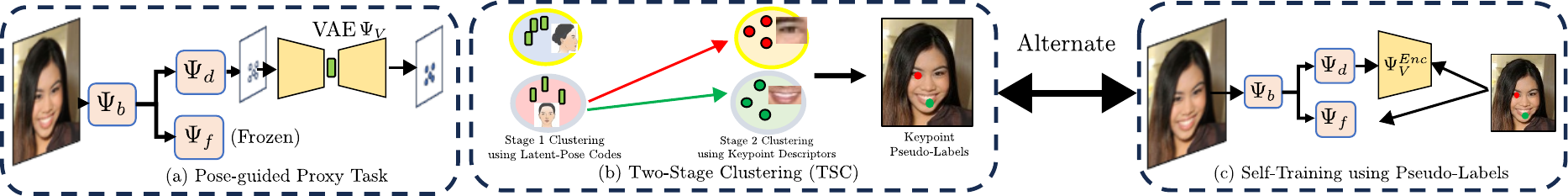}
    \caption{
    %The network architecture consists of a frozen Stable Diffusion module, an aggregator network $\Psi_b$, a descriptor  $\Psi_f$ and detector head $\Psi_d$.$\Psi_d$ outputs a keypoint heat-map.    
     Proposed diffusion based unsupervised landmark detection algorithm D-ULD++:  (a) Pose-guided proxy task to reduce noisy landmarks. (b) Two-stage clustering to improve pseudo-labels. (c) Self-training using pseudo-labels.}
     % For D-ULD, during training we alternate between computing keypoint correspondences via clustering and self-training using these  correspondences. With D-ULD++ we introduce a pseudo-pose learning proxy task to encode the keypoint spatial configuration into latent codes via a variational auto-encoder (VAE). We then perform a two-stage clustering. We cluster the latent codes into pseudo-pose clusters. Then for the images corresponding to the latent codes within each cluster, we further cluster the keypoints giving more fine-grained, pose specific keypoints which we use for self-training the network.}
    \label{fig:architecture_diagram}
    \vspace{-1.3em}
\end{figure*}

\section{Related Work}
\label{sec:related_work}

\noindent\textbf{Unsupervised Landmark Detection (ULD):} An early attempt at ULD involved enforcing equivariance constraints on a landmark detector. These constraints provide  self-supervision by requiring the features produced by a detector to be equivariant to the geometric transformations of an image. For ULD, equivariance constraints on image and pre-defined deformations were used along with auxiliary losses, such as locality \cite{thewlis2019unsupervised, thewlis2017unsupervised_dense}, diversity \cite{thewlis2017unsupervised} and others \cite{cheng2021equivariant}.  Albeit effective in discovering landmarks, such approaches struggle to produce semantically meaningful landmarks under large intra-class variations \cite{mallis2023keypoints}.

A promising class of methods proposed conditional image generation as a proxy task where the landmark detector is required to produce the geometry of an object \cite{jakab2018unsupervised, sanchez2019object, awan2023unsupervised}. 
%In this pipeline, a landmark detector is tasked to capture the geometry of an object, which is then used along with a deformed image by a decoder to reconstruct the original image. 
The two distinct components in such pipelines are: a  bottleneck that captures image geometry and a conditional image generator. % \cite{jakab2018unsupervised, sanchez2019object}. %The detector and bottleneck are responsible for producing object's geometry which is then combined with the deformed version of an input image and sent to a conditional image generator for reconstructing the original version of the input image. 
Zhang et al. \cite{zhang2018unsupervised} considered landmark discovery as an intermediate step of image representation learning and proposed  to predict landmark coordinates  utilizing soft constraints. Others combined image generation and equivariance constraints to obtain landmark representations \cite{cheng2021equivariant, kulkarni2019unsupervised, lorenz2019unsupervised}.
 %\cite{kulkarni2019unsupervised} introduced a keypoint bottleneck to transport learned image features between video frames; obtaining keypoints that consistently track objects and their parts.
%
Lorenz et al. \cite{lorenz2019unsupervised} disentangled shape and appearance  leveraging equivariance and invariance constraints. Wiles et al. \cite{wiles2018self} proposed a self-supervised learning for facial attributes from unlabeled videos which are then utilized to predict landmarks by training a linear layer on top of learned embedding. Most of these methods are prone to detecting semantically irrelevant landmarks under large pose variations.

\noindent\textbf{Clustering driven Self-Training:} In self-training methods, a model's own predictions are utilized as pseudo-labels (PLs) for generating a training signal. Typical approaches utilize highly confident predictions as hard PLs \cite{sohn2020fixmatch, xie2020self}, or via model ensembling \cite{nguyen2019self}. Self-training has been mostly leveraged for image classification~\cite{sohn2020fixmatch, xie2020self, rizve2020defense}, but also in other tasks~\cite{dai2015boxsup, khoreva2017simple,zhang2017supervision}.

Some other self-training methods use clustering to construct pseudo-labels \cite{asano2019self, caron2018deep, caron2020unsupervised, noroozi2018boosting, yan2020clusterfit, zhuang2019local}. Typically, the clustering serves to assign PLs to the unlabelled training images which are then used to create supervision signal \cite{caron2018deep, noroozi2016unsupervised}. Some other related methods are slot-attention mechanisms \cite{kipf2021conditional, locatello2020object}. 
A few mapped image patches into a categorical latent distribution of learnable embeddings \cite{razavi2019generating, van2017neural}, and proposed routing mechanisms based on soft-clustering \cite{lee2019set}. %The primary goal in these methods is to learn a powerful feature representation in an unsupervised way that can be applied to downstream tasks \cite{mallis2023keypoints}. 
In ULD, the work of Mallis et al.~\cite{mallis2020unsupervised, mallis2023keypoints} forms landmark correspondence through clustering landmark representations. This clustering is used to select pseudo-labels for first stage self-training. We also cluster landmark representations to generate PLs, however, we use it to propose a landmark detection method based on stable diffusion.% model.
\begin{figure*}
    
\end{figure*}
\noindent\textbf{Diffusion Model:} Diffusion models~\cite{song2019generative, song2020score,ho2020denoising,dhariwal2021diffusion}  generate better quality images on ImageNet~\cite{deng2009imagenet} compared to GANs. Recently, latent diffusion models~\cite{rombach2022high} facilitated their scaling to large scale data \cite{schuhmann2022laion}, also democratising high-resolution image synthesis by introducing the open-sourced text-to-image diffusion model, namely Stable Diffusion. Owing to its superlative generation capability, recent works  explore the internal representations of diffusion models \cite{tumanyan2023plug, hertz2022prompt, baranchuk2021label, xu2023open}. For instance, \cite{baranchuk2021label, xu2023open} investigate adapting pre-trained diffusion model for downstream recognition tasks. Different from these methods, we explore the potential of pre-trained Stable Diffusion for ULD.

% for first and second stages in the method. The pseudo-labels are then used to learn a landmark detector in a supervised manner in the second stage.

%by thresholding highly confident predictions and retaining confident predictions . 

%%%%%%%%% BODY TEXT
\section{Proposed Diffusion Based ULD Algorithm}
\label{sec:method}

Existing works targeting ULD struggle under  intra-class variations. In pursuit of overcoming these challenges, we explore the potential of pre-trained diffusion-based generative models \cite{ho2020denoising, song2020score,rombach2022high} for ULD.
%
%The success of diffusion models in generating high-quality images, suggests that their internal representations contain meaningful cues which could be useful toward solving tasks other than image synthesis. Some recent works 
%have used the representations of these pre-trained web-scale diffusion models like Stable Diffusion to achieve state-of-the-art performance in semantic correspondence tasks~\cite{luo2023diffusion, zhang2023tale, tang2023emergent}. 
We first perform a simple clustering of the pre-trained internal representations of the diffusion model, taken at random positions within object RoI. At inference, nearest neighbour querying is used to discover zero-shot landmarks (Sec.~\ref{subsection:Motivation: A Simple Clustering is all You Need}). %that outperforms existing works. 
Motivated by the performance of this zero-shot baseline, we propose an unsupervised landmark detection algorithm (D-ULD) founded on diffusion features (Sec.~\ref{subsec:Stable Diffusion based ULD pipeline}). 
%that uses self-training for pseudo-labels and clustering of keypoint descriptors for pseudo-labeling and emerges as a strong baseline (Sec.~\ref{subsec:Stable Diffusion based ULD pipeline}). %to learn correspondences across object images. 
We further improve D-ULD by introducing a new pose-guided proxy task (Sec.~\ref{subsubsection:Pose-guided Proxy Task}).
%harness the potential of this strong baseline by introducing a new pose-guided proxy task (Sec.~\ref{subsubsection:Pose-guided Proxy Task}). 
It reconstructs landmarks after projecting them to a latent pose space. % 3D pose as latent codes. 
To better capture the intra-class variations in landmark representations, we exploit this unsupervised latent pose space information to propose a two-stage clustering mechanism (Sec.~\ref{subsubsection:Two-stage Clustering mechanism}). After incorporating pose-guided proxy task and two-stage clustering mechanism in D-ULD, we contribute a new algorithm for unsupervised landmark discovery, dubbed as D-ULD++. ~\cref{fig:architecture_diagram} provides an overview of D-ULD and D-ULD++ algorithms.

\noindent\textbf{Problem statement:} We assume the availability of a set of images  $\SX = \{\mathbf{x} \in \RR^{H \times W \times 3}\}$ of a specific object category e.g., faces. After learning an initial set of keypoints on $\SX$ via SILK~\cite{gleize2023silk}, our training set becomes $\SX = \{\mathbf{x_j}, \{\mathbf{p_i^j}\}_{i=1}^{N_j} \}$, where $\mathbf{p}_i^j \in \RR^2$ is a keypoint in 2D space and $N_j$ is the number of keypoints detected in image $j$. These learned keypoints don't necessarily correspond to ground truth landmarks. 
%\textcolor{red}{More than one object landmark will not be part of $\mathbf{p_i^j}$}. 
%Keypoints may cover many areas in an image, and could also include outliers. 
Relying only on $\SX$, our aim is to train a model $\Psi: \SX \rightarrow \SY$, where $\SY \in \RR^{H^o \times W^o \times K}$ is the space of output heatmaps representing confidence maps for each of the $K$ object landmarks we wish to detect \cite{mallis2023keypoints}. We assume $[N]$ denotes the set $\{1, \hdots, N\}$.  

\noindent\textbf{Diffusion model overview:} Diffusion models are  generative models that approximate the data distribution by denoising a base data distribution (assumed Gaussian). In the forward diffusion process,  the input image \textbf{I} is gradually transformed into a Gaussian noise over a series of $T$ timesteps. %What is meant by Gaussian? Is it Gaussian noise?
Then a sequence of denoising iterations $\epsilon_{\theta}(\mathbf{I}_t,t)$, parameterized by $\theta$, and $t = 1 \hdots T$ take as input the noisy image $\mathbf{I}_t$ at each timestep and predict the noise $\epsilon$ added at that iteration \cite{ho2020denoising}. %The diffusion objective is given by {:
%\begin{equation}
%    \mathcal{L}_{DM} = \mathbb{E}_{\mathbf{I},t,\epsilon \sim \mathcal{N}(0,1)}  \left[ || \epsilon - \epsilon_\theta(\mathbf{I}_t,t) ||_2^2  \right].
%\end{equation}
Latent Diffusion models (LDM)~\cite{rombach2022high}, instead of operating on images directly encode images  as latent codes \textbf{z} and perform diffusion process. A decoder maps the latent representation to the image again.  %The objective for an LDM is:
\begin{equation}
    \mathcal{L}_{LDM} = \mathbb{E}_{\mathbf{I},t,\epsilon \sim \mathcal{N}(0,1)}  \left[ || \epsilon - \epsilon_\theta(\mathbf{z}_t,t) ||_2^2  \right].
\end{equation}
The denoiser for LDM consists of self- and cross-attention layers~\cite{vaswani2017attention}. We make use of the features of a pre-trained Stable Diffusion LDM for our method \cite{rombach2022high}. % \ie \textcolor{red}{we use Stable Diffusion as frozen}. %This is not required.
\subsection{Proposed Zero-Shot Baseline}
\label{subsection:Motivation: A Simple Clustering is all You Need}
%\noindent\textbf{Motivation:} %Recent works ~\cite{luo2023diffusion, zhang2023tale, tang2023emergent} have shown the internal representations of diffusion models e.g., stable diffusion are useful for tasks beyond image synthesis, such as semantic correspondence. 
%It has recently been found that  features from diffusion models are effective for the task of semantic correspondence~\cite{luo2023diffusion, zhang2023tale, tang2023emergent}.  It signifies that these generative models understand the underlying semantics of common objects. Motivated by this, 
\noindent We propose ZeroShot, a zero-shot baseline to measure the efficacy of diffusion features for the ULD task, comprised of feature aggregation, clustering and exemplar assignment. %Below, we describe the methodology for our zero-shot baseline.

%Motivated by this, we first developed a simple baseline to measure the efficacy of these diffusion features for the task of unsupervised landmark detection. We use a frozen Stable Diffusion models as our feature extractor. 

\noindent\textbf{Feature aggregation:} Feature maps in Stable Diffusion are spread over network layers and diffusion time-steps. So, extracting useful feature descriptors from them is a non-trivial task. Similar to~\cite{luo2023dhf}, we  employ a network that aggregates features over layers and time-steps to obtain a  single feature map.
The aggregator network, takes as input the features maps $\mathbf{r}_{l,t}$  obtained from Stable Diffusion ($\mathrm{SD}$) for an input image $\mathbf{x}_j$. We upscale $\mathbf{r}_{l,t}$ to image resolution and pass it through a bottleneck layer $\mathbf{B}_l$~\cite{he2016deep,xu2023open} to obtain a fixed channel count, and weight it with a mixing weight $\mathbf{w}$. The final aggregated feature map is: %given by: %confirm the lefthand side
\begin{equation}
\small
 {\mathbf{F}_{ag}=}   \sum_{t=1}^{T} \sum_{l=1}^{L} \mathbf{w}_{l,t}\mathbf{B}_l(\mathbf{r}_{l,t}), \{\mathbf{r}_{l,t}\}_{l=1,t=1}^{L,T}=\mathrm{SD}(\mathbf{x})
    \label{eqn:feat-aggr}
\end{equation}
where $L$ is the number of layers, $T$ is the number of timesteps. %$l$ and $t$ are layer and time indices, respectively. 
%Bottleneck layers are shared across timesteps %giving $L$ in total 
%to maintain a constant feature dimension. %across layers.
There are $L \times T$  combinations of layer indices and timesteps. We learn unique mixing weights $\mathbf{w}_{l,t}$ across all layer and timestep combinations via backpropagation. %We denote the aggregator network as $\Psi_{b}$ which is a pre-trained network of~\cite{luo2023dhf}. 
The overall process of getting feature map $\mathbf{F}_{ag}$ may be defined as: $\mathbf{F}_{ag}=\Psi_b(\mathbf{x})$, where $\Psi_b$ includes both the stable diffusion process and the feature aggregation. %We use the pre-trained weights from~\cite{luo2023dhf} trained for semantic correspondence. %The output feature map from $\Psi_{b}$ have $128$ channels. %From henceforth on, when referring to feature extraction, we specifically mean extraction from the aggregator network. %loss function of aggregator?

\noindent\textbf{Clustering:} 
For every image in the training set, we randomly sample pixels within the ROI region in the output feature map $\mathbf{F}_{ag}$. For each sampled pixel, we get a descriptor which is then clustered using K-means and the cluster centroids from the training set are retained. Detectron2~\cite{wu2019detectron2} is employed to detect the ROI.
%The first step toward constructing this baseline involves randomly sampling pixels within the image ROI (e.g., human and cat faces), and extracting their corresponding descriptors from $\Psi_b$. %Why randomly sampling? Why not key points referred previously?
%Such features are extracted for every image of the training set. In the second step, they are clustered via K-Means and the cluster centroids from the training set are retained. 
On the test set, again random pixels are sampled for each image within ROI, and after their corresponding descriptors are extracted, they are assigned the label of their closest  cluster. Each image may then consist of multiple pixel locations belonging to a particular cluster. %Each cluster can thus have multiple keypoints associated with it per image. 
As some of these can be noisy assignments, we prune the locations so that per image each cluster has a single best location via {exemplar assignment}. 

\noindent\textbf{Exemplar assignment:} We assign to each image at most $K$ pixel locations by choosing for each cluster the locations whose descriptor is closest to the cluster centroid in feature space. We discard the remaining locations, leaving for each image at most $K$ pixel locations. These $K$ pixel locations after assignment are considered as discovered landmarks. 

%(some clusters may have no keypoints in an image). %How zero assignment is decided? Threshold? 
%\textcolor{red}{Since this set of operations is referenced multiple times, we opt for the shorthand \emph{exemplar assignment} to simplify communication.}
% This sentence is not required. 
%  refer to this set of operations multiple time 
%The exemplar assignment completes our zero-shot baseline  dubbed as ZeroShot.
%method is referred to as ZeroShot in our experiments
% On the test set, we randomly sample keypoints for each image, and after extracting their corresponding descriptors, we assign them label by finding  assigned 
% extracted and assigned the label of their closest corresponding cluster. 
% This method is referred to as ZeroShot in our experiments. 

\noindent\textbf{Discussion:} %\Cref{tab:results} shows the results of ZeroShot on the four different datasets. \emph{This simple baseline outperforms previous approaches without any training by margins of $7-10\%$ across all four datasets.} 
\cref{fig:method-qual} shows some qualitative results of ZeroShot on the AFLW dataset. While ZeroShot is able to reliably detect keypoints for most front-oriented faces, it occasionally tends to confuse keypoints on the left and right side of side-oriented faces for some examples. Further, it tends to have poor keypoint localization on side profile faces. 
\cref{fig:method-quant} shows the forward and backward NME\% for ZeroShot  amongst other methods for various yaw angles on the AFLW dataset. Errors for the front-facing angles $(-30^{\circ} \text{ to } 30^{\circ})$ are significantly lower than those of the more side-oriented ones. Compared to \cite{mallis2023keypoints}, an error reduction of $14.6\%$ in  front-oriented faces is seen. Whereas a more modest improvement of $4.5\%$ is seen for side-oriented faces (yaw angles $(-60^\circ \text{ to }-30^\circ)$ and $(30^\circ \text{ to }60^\circ)$) and $2.1\%$ for more extreme view points (yaw angles $(-90^\circ \text{ to }-60^\circ)$ and $(60^\circ \text{ to }90^\circ)$). %We show additional analysis of ZeroShot in~\Cref{sec:experiments} on the LS3D Dataset. %We also provide additional per-keypoint localization accuracy analysis for ZeroShot in the supplementary. 
%
%To measure per-keypoint localization accuracy, we make use of the ground-truth keypoints provided by the LS3D dataset. This dataset consists of face images, each annotated with 68 keypoints. We take from a training set image $\mathbf{x}_j$, a ground-truth keypoint $\mathbf{p}^j_i$. In a test image $\mathbf{x}_{j'}$, we find it's corresponding keypoint $\mathbf{p}^{j'}_i$ by a nearest neighbor assignment in the feature space between the descriptor of $\mathbf{p}^j_i$ and the  descriptors of the keypoints of $\mathbf{x}_{j'}$ . We then compute the pixel distance between $\mathbf{p}^{j'}_i$ and the ground truth keypoint in $\mathbf{x}_{j'}$ corresponding to $\mathbf{p}^j_i$, denoted as $\overline{\mathbf{p}}^{j'}_i$. This distance $||\mathbf{p}^{j'}_i-\overline{\mathbf{p}}^{j'}_i||$ is measures how accurately the keypoints are localized.
%Using the keypoints obtained in the test set images by ZeroShot
The limitations of ZeroShot motivate our next contribution.% where we propose an unsupervised landmark detection (ULD) algorithm based on diffusion features (D-ULD); it leverages pseudo-labeling of keypoints via descriptor clustering and self-training with the constructed pseudo-labels.

%propose to iteratively alternate
%between pseudo-labelling of keypoints  via descriptor clustering, and
%model self-training with the produced pseudo-labels. 

%\subsection{Keypoint Self-Training} 

\subsection{Proposed  Diffusion-based D-ULD Algorithm} 
\label{subsec:Stable Diffusion based ULD pipeline}

\noindent In this section, we describe our unsupervised landmark detection algorithm (D-ULD) based on diffusion features that uses clustering to obtain pseudo-labels (PLs) and employs these PLs for self-training.

\noindent\textbf{Network Structure:} We append a descriptor head $\Psi_f$  and a landmark detector head $\Psi_d$ after $\Psi_b$. 
%Using the feature map $F_{ag}$ obtained from $\Psi_b$, we append  a descriptor $\Psi_f$ and a detector $\Psi_d$. 
%
The detector head $\Psi_d$ will produce a single-channel heatmap $\mathbf{H}_j = \Psi_d(\Psi_b(\mathbf{x}_j)) \in \mathbb{R}^{H \times W \times 1}$ for image $\mathbf{x}_j$. This heatmap reveals the confidence of the model for an object landmark at a given location. Non-maximum suppression is used to obtain landmarks from $\mathbf{H}_j$ \cite{mallis2023keypoints}. The descriptor $\Psi_f$ produces for image $\mathbf{x}_j$ a descriptor volume $\mathbf{F}_j = \Psi_f(\Psi_b(\mathbf{x}_j)) \in \mathbb{R}^{H \times W \times D}$, where $D$ is the feature dimension. From $\mathbf{F}_j$, we extract a feature descriptor $f_i^j \in \RR^D$ corresponding to the landmark location found by $\Psi_d$.  We denote the network consisting of $\Psi_b$, $\Psi_f$ and $\Psi_d$  as $\Psi$.

\noindent\textbf{Bootstrapping:} We initially train the network $\Psi$ (except the stable diffusion network) via a self-supervised keypoint training scheme~\cite{gleize2023silk}. %Our objective is to improve on the random pixel locations used in ZeroShot baseline, by learning dataset specific keypoints.
Specifically, during bootstrapping the network $\Psi$ is trained by learning correspondences between an image and it's augmentation, by minimizing a BCE loss for the detector head $\Psi_d$ and a negative-log-likelihood loss for the descriptor head $\Psi_f$. We use flips and random rotations as augmentations. After the initial keypoint learning, 
the training set consists of images, learned keypoints and the corresponding descriptors, \emph{i.e.} $\SX = \{\mathbf{x}_j, \{\mathbf{p}^j_i,\mathbf{f}^j_i\} \mid i \in \text{keypoints for image} \ \mathbf{x}_j  \}$.

\noindent\textbf{Self-Training:}
To improve landmark detection across a larger spectrum of viewpoint changes and symmetric view pairs, we resort to a self-training scheme that uses keypoint pseudo-labelling \cite{mallis2020unsupervised}.
%
%use a keypoint pseudo-label based self-training scheme.
%
We initialize this step by clustering all keypoint descriptors $\{ \mathbf{f}^j_i \mid \forall i \in \text{keypoints in } \mathbf{x}_j, \forall j \in \text{images}\}$ into $K$ clusters using $K$-means. Then, we perform exemplar assignment and remove redundant keypoints, leaving for each image $\mathbf{x}_j$ a set of $K$ keypoints and their corresponding descriptors $\{\mathbf{p}^j_i,\mathbf{f}^j_i \mid \forall i \in [K] \}$. This provides us with a pseudo-label $\mathbf{c}^j_i  \in [K]$ for keypoint $\mathbf{p}^j_i$. % $\mathbf{c}^j_i \in [K]$ denotes the pseudo-label for $\mathbf{p}^j_i$.
%
 %we also remove redundant keypoints for image $x_j$ by only retaining a single descriptor per cluster for each image. For a given image $j$, we find for each
%cluster $k$, from keypoints $\{\mathbf{p}^j_i \mid \forall i \in \text{image j}\}$ the 
% descriptor closest to the centroid and remove duplicate occurrences of the same
%cluster label $k$. Thus for each image $\mathbf{x}_j$ there are a set of $\leq K$ keypoints as some clusters may not occur in an image. 
%
Updating the training dataset with pseudo-labels, our dataset is now $\SX = \{\mathbf{x}_j, \{\mathbf{p}^j_i,\mathbf{f}^j_i,\mathbf{c}^j_i  \}_{i=1}^{K} \}$. 

%$\mathbf{c}^j_i \in \{1, \hdots K\}$ denotes the pseudo-label for $\mathbf{p}^j_i$. %Without loss of generalization we assume all clusters are present while missing clusters for an image  will  be ignored without changing the formulation.
%
%Keypoint pruning also provides us keypoint correspondences across images of the training set, as for each $\mathbf{p}^j_i$ for image $i$, there exists at most one keypoint $\mathbf{p}^{j'}_{i'}$ in image $j'$ with the same cluster label, \ie $\mathbf{c}^{j}_{i} = \mathbf{c}^{j'}_{i'}$.
%We exploit these cluster labels as landmark pseudo-label correspondences in our self-training scheme. 

%
%As every cluster has a representative keypoint in each image, this also provides correspondences across images which we intend to exploit for our self-training scheme.

% \begin{figure}
%     \centering
    
%     \caption{Comparison of Forward and Backward errors (as NME \%) split by yaw-angles ranges on the AFLW~\cite{koestinger2011annotated} dataset for our method, the Zero Shot baseline with comparison to Mallis~\cite{mallis2023keypoints}.}
%     \label{fig:method_nme}
% \end{figure}

\begin{figure}
  \centering
    \subfloat[]{
    \centering
    \includegraphics[width=0.45\textwidth]{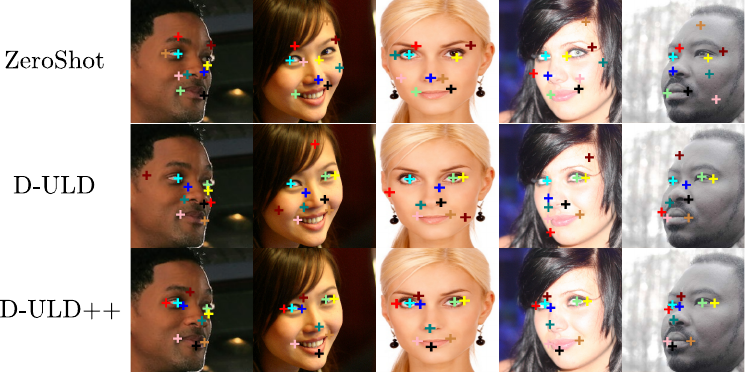}
    \label{fig:method-qual}
  }\newline
  \subfloat[]{\includegraphics[width=0.40\textwidth]{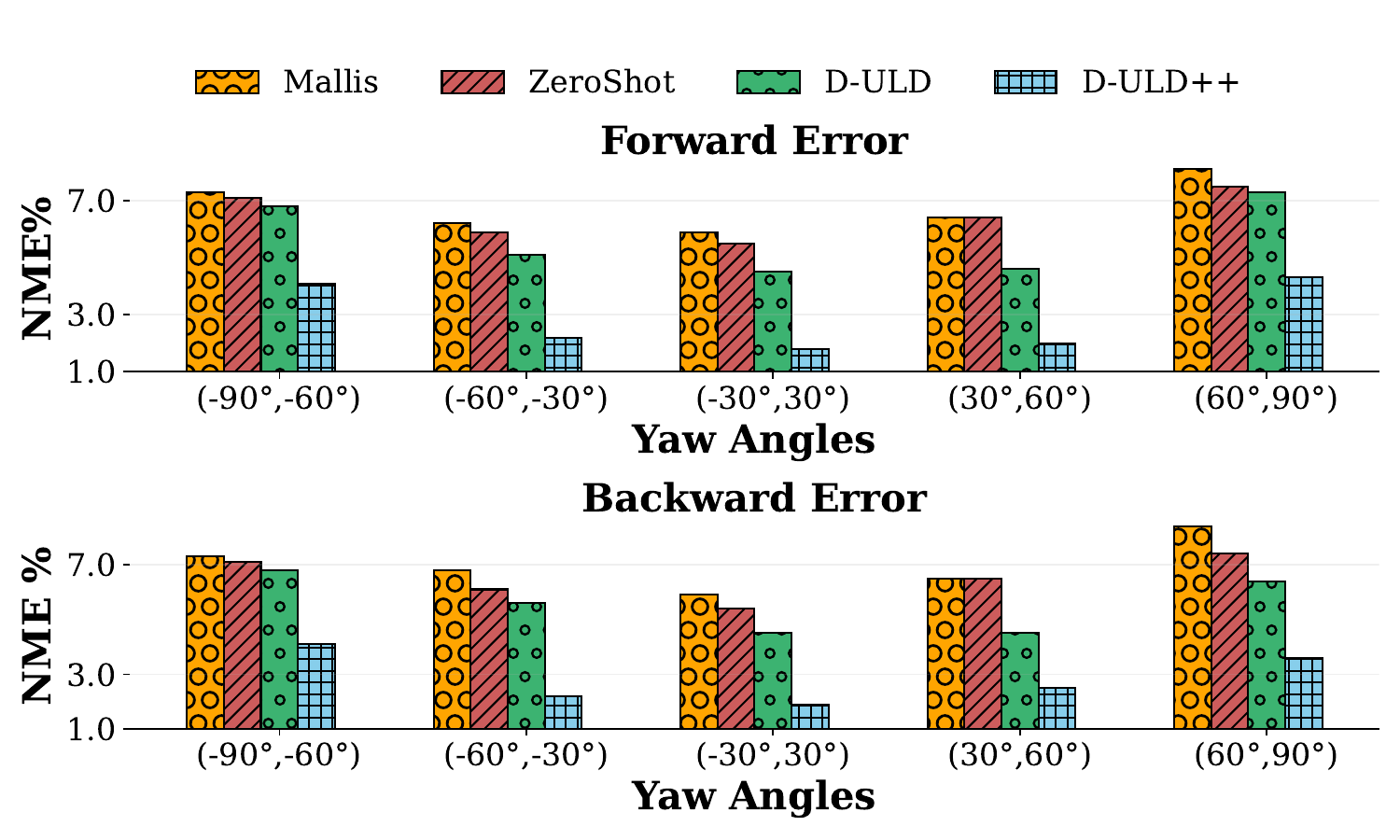}
    \label{fig:method-quant}}
  \caption{Comparisons of ZeroShot, D-ULD and D-ULD++. (a) Visual results on exemplar images showing different colored keypoints. 
  (b) Yaw angle split of fwd.~and bwd.~errors  (NME\%) for AFLW dataset. Mallis \cite{mallis2023keypoints} is shown for additional comparison.}
  % \label{fig:method-qual} 
  \vspace{-1em}
\end{figure}
\noindent\textbf{Training objectives:} We use the pseudo-labels $\mathbf{c}^j_i$ contained in the training set $\SX$ to train the network $\Psi$.
%After the initial clustering and image keypoint assignment step, each image $\mathbf{x}_j$ has associated with it a set of keypoints $\{ \mathbf{p}^j_i \mid j \in \text{images}, \ i \in  \text{keypoints for image} \ j \}$ and pseudo-labels $\mathbf{y}_j$. 
%The network $\Psi$ consisting of the aggregation network $\Psi_b$, detector $\Psi_d$ and descriptor heads $\Psi_f$, is now trained using these pseudo-labels. 
The loss corresponding to the detector head is the mean-squared error (MSE) loss between the pseudo-label heatmaps and detector head output. %Here the heatmap $G$ for a given image $\mathbf{x}_j$ is formed by placing 2D-Gaussian maps on each of
To train the descriptor head $\Psi_f$, we use a contrastive loss that pulls feature descriptors  $\mathbf{f}^j_i$ from the same cluster together and pushes features from different clusters away from each other \cite{mallis2023keypoints}.
Once the training is done for a fixed number of epochs, the training set is updated with improved keypoints, descriptors, and new pseudo-labels. Thus, $\mathbf{p}$, $\mathbf{f}$ and $\mathbf{c}$ are functions of the epoch $t$. We refer to the training set at epoch $t$ as $\SX_t = \{\mathbf{x}_j, \{\mathbf{p}^j_i(t), \mathbf{f}^j_i(t), \mathbf{c}^j_i(t) \}_{i=1}^K \}$.
%

% :
% \begin{equation}
%     \SL_{d} = ||G(\mathbf{x}_j) - \Psi_d(\Psi_b(\mathbf{x}_j)) ||^2
%     \label{eqn:mse-detector}
% \end{equation}

% Here the ground-truth heatmap $G$ for a given image $\mathbf{x}_j$ is formed by placing 2D-Gaussian maps on each of
% the keypoint locations.
%
% To train the descriptor head $\Psi_f$, we use a contrastive loss that pushes features from the same cluster together. Let $\mathbf{p}^j_i$ and $\mathbf{p}^{j'}_{i'}$ be two keypoints indexed by $i$ and $i'$ for images $j$ and $j'$ respectively. The loss we minimize to train $\Psi_f$ is:

% \begin{eqnarray}
%     \SL_f(\mathbf{x}_i^j,\mathbf{x}_{i'}^{j'}) = \mathbf{1}_{[\mathbf{c}^{j'}_{i'} = \mathbf{c}^{j}_{i}]} ||\mathbf{f}^j_i - \mathbf{f}^{j'}_{i'} || + \nonumber \\ \mathbf{1}_{[\mathbf{c}^{j'}_{i'} \neq \mathbf{c}^{j}_{i}]}\text{max}(0,m-||\mathbf{f}^j_i-\mathbf{f}^{j'}_{i'}|| \label{eq:desc-loss}
% \end{eqnarray}

% where recall $\mathbf{f}^j_i = \Psi_f(\Psi_b(\mathbf{x}_j))$ is the feature descriptor extracted at image $j$ for keypoint $\mathbf{p}^j_i$,. A margin $m$ is used to enforce
% features corresponding to negative pairs to be far apart.  Our choice of positive pairs is obvious, as $\mathbf{c}^{j'}_{i'} = \mathbf{c}^{j}_{i}$ for only a single pair of keypoints in any two images. For negative pair selection, for a particular keypoint location $\mathbf{p}^j_i$ we pick $\mathbf{p}^{j'}_{i'}$ to be the farthest point by pixel distance for which $\mathbf{c}^{j'}_{i'} \neq \mathbf{c}^{j}_{i}$ holds.

\noindent\textbf{Discussion:} \cref{fig:method-quant} shows that our D-ULD algorithm improves significantly over ZeroShot across all yaw-angles, %Especially in the yaw-angle ranges $(-60^{\circ},-30^{\circ})$ and $(30^{\circ},60^{\circ})$ D-ULD improves upon ZeroShot by $\mathbf{XXX}$. 
In~\cref{fig:method-qual} for D-ULD, landmarks on the eye pupils, nose and mouth are accurately localized while the landmarks around the eyes require further improvement. %For example, the blue landmark  around the eye region in the center two images and on the forehead region in the two side images. 

\subsection{Proposed D-ULD++ Algorithm} 
\label{subsec:D-ULD++}

While D-ULD algorithm improves upon the ZeroShot baseline across pose variations, it still has a tendency to output semantically non-meaningful landmarks for extreme poses.  It is because the landmark descriptors inherently contain local image information, and there is no explicit mechanism in the D-ULD to capture the collective spatial configuration of landmarks. To this end, we design a simple proxy task that takes the predicted landmark heatmaps from D-ULD and aims to reconstruct them after projecting into a low dimensional latent pose space. Next, to better capture the variations caused by large viewpoint changes, we propose a two-stage clustering mechanism.

\subsubsection{Pose-guided Proxy Task}
\label{subsubsection:Pose-guided Proxy Task}
% While our ULD pipeline is capable of improving the landmark detection across pose variations, it still has a tendency to learn semantically non-meaningful landmarks for more extreme variations.  This is because keypoints inherently contain local information image information. To handle these extreme pose variations, we propose to learn an image level representation of these learned keypoints via a proxy task, that we subsequently use to inject some pseudo-pose information via a two-stage clustering scheme~\Cref{sec:two-stage}.
%Previous works have explored using the set of predicted keypoints from a network to detect poses~\cite{} for human pose estimation. We propose a similar proxy task, to learn image level representations from the predicted keypoints. 

A relationship between the  facial landmarks and the corresponding pose of that face is quite intuitive. For example, the visibility of particular landmarks can be easily associated with a particular pose. Thus each pose constraints the landmark visibility and location to a relatively smaller space while each set of landmarks constraint the facial pose as well. Given accurate pose and landmark positions, a mapping may be learned in both directions. However, in our case, the pose supervision  is not available while the landmark positions are noisy. We therefore propose a proxy task to project the noisy landmarks to a latent pose space and then back project to the original space with the aim of removing the localization noise in landmarks.  
This proxy task constrains the spatial configuration of landmarks with respect to each other.

%Our goal is to design a proxy task that is tightly connected to the task of landmark detection. In other words, while communicating with the landmark detector, it can encode the landmark configuration, possibly exhibiting several variations, as a shape information of a deformable object. In pursuit of this, we map the predicted landmarks to a latent pose code embedding and then reconstruct the original predicted landmarks from this latent code embedding. We realize this as a variational auto-encoder (VAE) denoted as $\Psi_V$. This $\Psi_V$ is appended to our descriptor head $\Psi_d$.

%To this end, we  append a variational auto-encoder (VAE) denoted as $\Psi_V$ to our  descriptor head $\Psi_d$.
%Assuming the input to the network $\Psi$ is image $\mathbf{x}_j$ and the predicted output heat map by the detector head $\Psi_d$ is $\mathbf{H}_f^j$, \ie $\mathbf{H}_f^j = \Psi_d(\Psi_b(\mathbf{x}_j))$.
% 

For this purpose we append an variational autoencoder (VAE) $\Psi_V$ to the detector head $\Psi_d$. The input to $\Psi_V$ is the predicted landmark heatmap $\mathbf{H}_j$. $\Psi_V$ is trained to construct a pseudo-ground truth heatmap $\mathbf{G}_j$ formed by placing $K$ 2D-Gaussians corresponding to predicted landmark locations $\{ \mathbf{p}^j_i \mid i \in [K]\}$ on a single channel.
%
%on a single channel on each of the predicted landmark locations $\{ \mathbf{p}^j_i \mid i \in [K]\}$. 
%
The output of the encoder produces a latent pose code $\varphi_j$ corresponding to $\mathbf{H}_j$. 
%as the latent code $\mathbf{l}_j \in \mathbb{R}^{3}$.
This latent code is then used by the decoder to reconstruct the $\mathbf{G}_j$. %The $\Psi_V$ consists of residual blocks with \texttt{BatchNorm}-\texttt{LeakyRelu}-\texttt{Conv} blocks for the encoder and decoder. 
The dimension of the latent code vector $\varphi_j$ is fixed to 64. %The training loss for the this encoder-decoder architecture is the standard ELBO loss~\cite{kingma2013auto}. 
 
\noindent\textbf{Discussion:} The $\Psi_V$ is trained on only landmark information without any pose supervision. We observe in~\Cref{sec:experiments}, in-spite this limitation, the latent codes obtained after training in-fact capture relevant pose information. As such, a k-Means clustering of the latent codes is performed and the clusters thus obtained contain images corresponding to the same pose-range. The additional supervision by  $\Psi_V$ improves both forward and backward NME error by $7.8\% $ and $12.4\%$ on AFLW and $4.9\% $ and  $6.4\%$ on MAFL in~\Cref{tab:ablation}. %The more modest improvements in MAFL, in comparison to AFLW, can be attributed to more front-facing images in the dataset.
 %We leverage this `pseudo-pose' in~\cref{subsubsection:Two-stage Clustering mechanism} to learn keypoints consistent across wider pose variations. 
 
 %improve improve matching across images with greater pose variation, in the process improving  provide additional supervision for the network.

%We propose as a proxy task training our a variational auto-encoder, to reconstruction the keypoint heat map predicted by $\Psi_d$. Once trained, we show the latent codes learned by the VAEs contain pose information in~\Cref{sec:results}. 

 %The VAE is trained via a standard ELBO loss~\cite{kingma2013auto}. 
 %Once trained, the VAE can predict keypoint heatmaps. 
 
 %We show these predicted keypoint heatmaps give pose information. We show through analysis in the results section (\Cref{fig:aaa}) that the latent codes indeed represent poses by performing a K-Means clustering of the latent codes and the majority of the poses contained in the clusters are similar.

\subsubsection{Two-stage Clustering} 
\label{subsubsection:Two-stage Clustering mechanism}
In realistic settings, deformable objects like faces can undergo large variations like extreme 3D rotations. A simple clustering of keypoint descriptors is restrictive in capturing such variations due to the inherently local nature of keypoint information \cite{mallis2023keypoints, awan2023unsupervised}. This poses a limitation in achieving a robust ULD method.
%This inherent reliance on local information could present itself as a limitation in realizing a robust ULD pipeline.
To overcome this, we devise a two-stage clustering (TSC) mechanism which exploits the pseudo-pose generated by our pose-guided proxy task. In particular, TSC leverages the latent code to  perform  stage-1 pose-based clustering. Next, in each pose-based cluster, we perform stage-2 clustering of keypoint descriptors from images that belong to the same stage-1 cluster. Thus, TSC essentially decomposes the problem of recovering landmark correspondence under large intra-class variations into two stages. Moreover, it allows us to leverage both local and image-level information by using the learned keypoint descriptors and latent codes. Such a scheme facilitates reduced within-cluster variations compared to the simple clustering. 

% To improve landmark detections across wider pose variations we leverage both local and image-level information by using the learned keypoint descriptors and latent codes for self-training. 

In TSC, in the first stage, latent codes $\boldsymbol{\varphi}_j$ are partitioned into $Q$ clusters using k-means, and latent code labels $\boldsymbol{u}_j \in [Q]$ are assigned. In the second stage, we perform a further K-means clustering of a collection of keypoint descriptors $\mathbf{f}$ from images belonging to the same pose-based clusters. The number of clusters in the second stage is the same as the number of landmarks $K$ to be detected in a particular dataset. Thus, the two-stage clustering provides us with a total of $Q \times K$ clusters. $[R] = [Q] \times [K]$ denotes the new label set.
With this TSC, for given image $\mathbf{x}_j$, we find the pseudo-label $\mathbf{c}^{j}_{i}$, by retaining for each cluster $r \in [R]$, the keypoint whose descriptor is nearest to the cluster centroid. We then update the training set with the latent codes $\boldsymbol{\varphi}_j$, their labels $\boldsymbol{u}_j$, as well as the more fine-grained keypoint labels $\mathbf{c}^j_i$ obtained from TSC.
%the more fine-grained labels $\mathbf{c}^j_i$ generated from TSC, as well as the pose-labels obtained by clustering the latent codes denoted by $\{ \boldsymbol{u}_j \in [Q] \}$.
$\SX=\{\mathbf{x}_j, \{\mathbf{p}^j_i,\mathbf{f}^j_i,\mathbf{c}^j_i  \}_{i \in [R]}, \boldsymbol{\varphi}_j, \boldsymbol{u}_j  \}$ is the updated dataset. Now, similar to D-ULD, we alternate between the clustering and using the resulting pseudo-labels for supervision.

% add to the training set, the latent code pseudo-labels to the training set. Thus the training dataset now consists of $\SX = \{\mathbf{x_j}, \{\mathbf{p}^j_i, \mathbf{f}^j_i, \mathbf{c}^{j}_i\}, \boldsymbol{l}_j, \boldsymbol{u}_j \mid j \in \text{keypoints for image } j, j \in \text{images}  \}$, where $\boldsymbol{l}_j$ is the latent code corresponding to $\mathbf{x}_j$ and $\boldsymbol{u}_j$ is it's pose pseudo-label.
%
%Similar to D-ULD, we alternate between clustering and using the resulting pseudo-labels for supervision. However, now 

%However, now in addition to the keypoint pseudo-labels we also do so for the pose pseudo-labels.

\noindent\textbf{Training objectives:} D-ULD++ is trained in two steps. In first step, we backprop through $\Psi_v$ using ELBO loss. Similar to D-ULD, the detector head $\Psi_d$ is trained via the MSE loss with the pseudo-labels obtained using TSC, and the the descriptor head $\Psi_f$ is trained using the contrastive loss on keypoint descriptors $\mathbf{f}_i^{j}$. In the second step, we do not use the decoder and so the ELBO loss. The detector head is trained same as first step, while the descriptor head is kept frozen. Instead, we just use the encoder in $\Psi_v$ and backprop through it using the contrastive loss which encourages latent codes with the same labels to be close to each other: 
\begin{align}
\SL_d(\boldsymbol{\varphi}_j,\boldsymbol{\varphi}_{j'}) = \mathbf{1}_{[\boldsymbol{u}_j = \boldsymbol{u}_{j'}]} ||\boldsymbol{\varphi}_j - \boldsymbol{\varphi}_{j'} || + \nonumber \\ \mathbf{1}_{[\boldsymbol{u}_j \neq \boldsymbol{u}_{j''}]}\text{max}(0,m-||\boldsymbol{\varphi}_j-\boldsymbol{\varphi}_{j''}||)
\label{eqn:latent-opt}
\end{align}
where $\boldsymbol{\varphi}_j$ is the output of the VAE Encoder $\Psi_V^{Enc}$, \ie $\boldsymbol{\varphi}_j = \Psi_V^{Enc}(\Psi_d(\Psi_b(\mathbf{x}_j))$. 
We select any $\boldsymbol{\varphi}_{j'}$ and $\boldsymbol{\varphi}_{j''}$ such that $\boldsymbol{u}_j = \boldsymbol{u}_{j'}$ and $\boldsymbol{u}_j \neq \boldsymbol{u}_{j''}$ respectively. We provide a pseudo-code for D-ULD++ in the supplementary.

% First, we train $\Psi_v$ using ELBO loss, next we only backprop through the encoder with our contrastive loss formulation. The detector head $\Psi_d$ is trained via the MSE loss with the pseudo-labels obtained using TSC. To train the descriptor head $\Psi_f$  contrastive loss is used encouraging latent codes with the same labels to be close to each other. We minimize the following loss:
% \begin{align}
% \SL_d(\boldsymbol{\varphi}_j,\boldsymbol{\varphi}_{j'}) = \mathbf{1}_{[\boldsymbol{u}_j = \boldsymbol{u}_{j'}]} ||\boldsymbol{\varphi}_j - \boldsymbol{\varphi}_{j'} || + \nonumber \\ \mathbf{1}_{[\boldsymbol{u}_j \neq \boldsymbol{u}_{j''}]}\text{max}(0,m-||\boldsymbol{\varphi}_j-\boldsymbol{\varphi}_{j''}||)
% \label{eqn:latent-opt}
% \end{align}
% 

%As in D-ULD, at a fixed number of epochs, the training set is refreshed with the network's output, including new keypoints, descriptors, cluster assignments, and latent code embeddings with their respective cluster assignments.

\begin{figure*}
    \centering
    \includegraphics[scale=0.5]{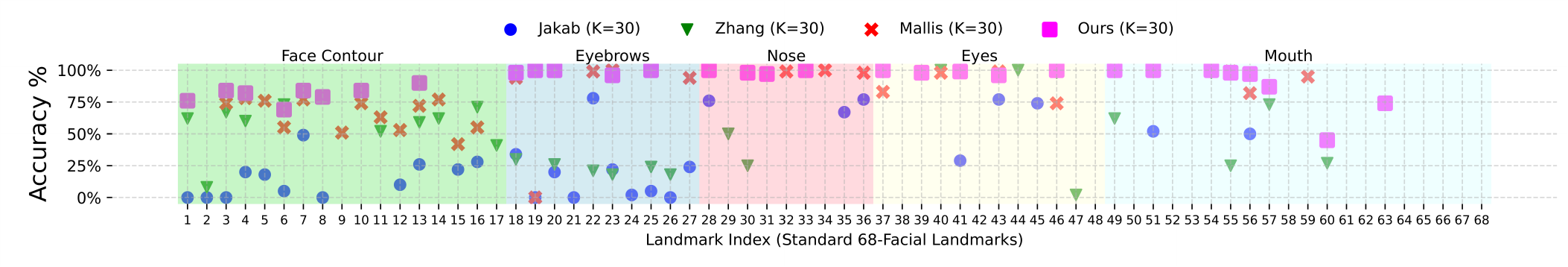}
    \vspace{-1em}
    \caption{Evaluation of the ability of raw unsupervised landmarks to capture supervised landmark locations on MAFL. Each unsupervised landmark is mapped to the best corresponding supervised landmark using the Hungarian Algorithm. Then accuracy is calculated for a distance threshold of $0.2\cdot$ $d_{iod}$ to a landmark location, where $d_{iod}$ is the interocular distance. Accuracy is shown for each of the 68-facial landmarks sorted by ascending order of index. Different landmark areas are highlighted with different colours and labelled as such (1-17 face contour, 18-27 eyebrows etc.}
    \label{fig:hungarian-plot}
    \vspace{-1em}
\end{figure*}

\begin{figure*}
    \centering
  \subfloat{
  \centering
    \includegraphics[width=0.95\textwidth]{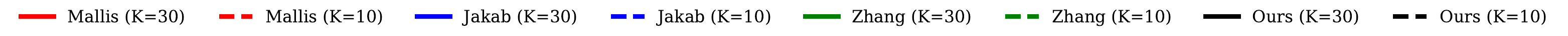}
    \label{fig:legend-ced}}\newline
  \subfloat{
    \includegraphics[width=0.21\textwidth]{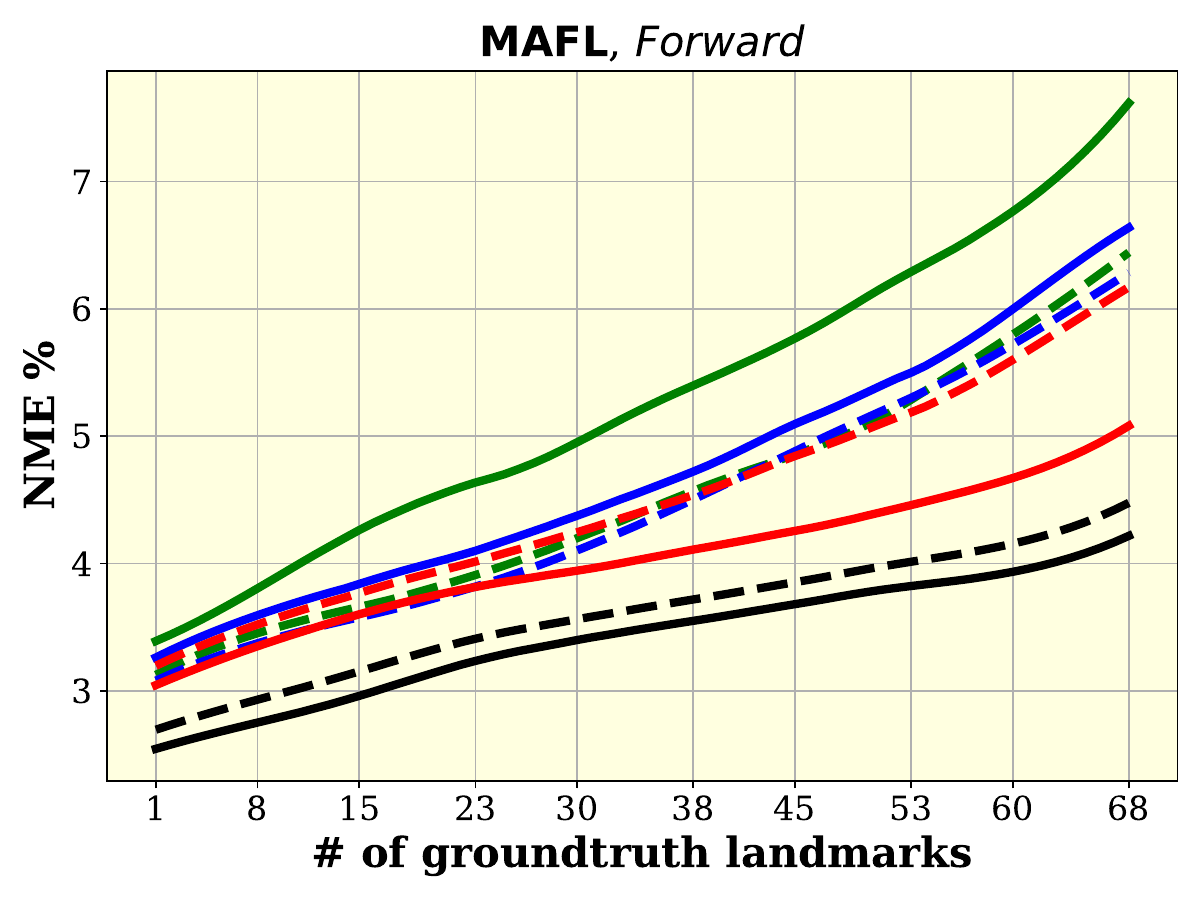}
    \label{fig:mafl-fwd}}
  \subfloat{
    \includegraphics[width=0.21\textwidth]{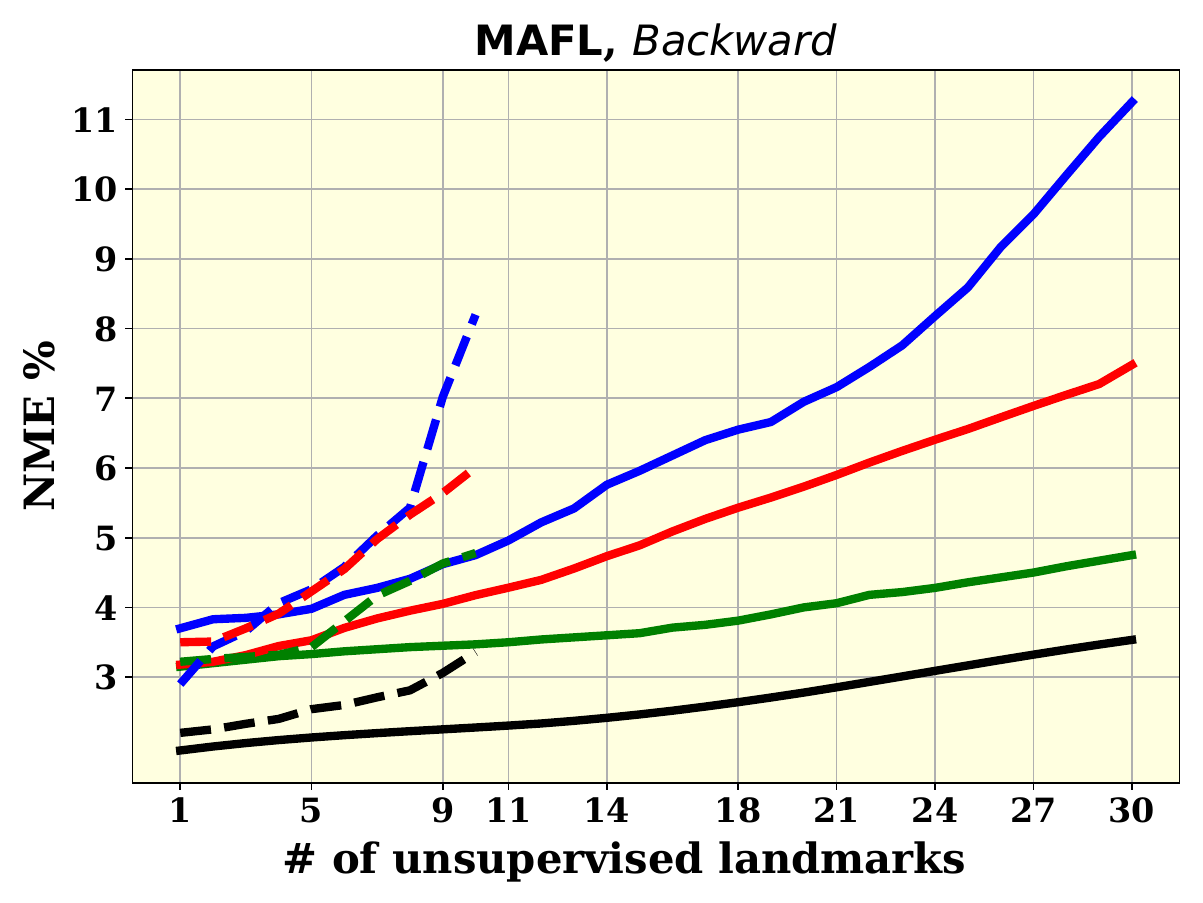}
    \label{fig:mafl-bwd}
  }
  \subfloat{
    \includegraphics[width=0.21\textwidth]{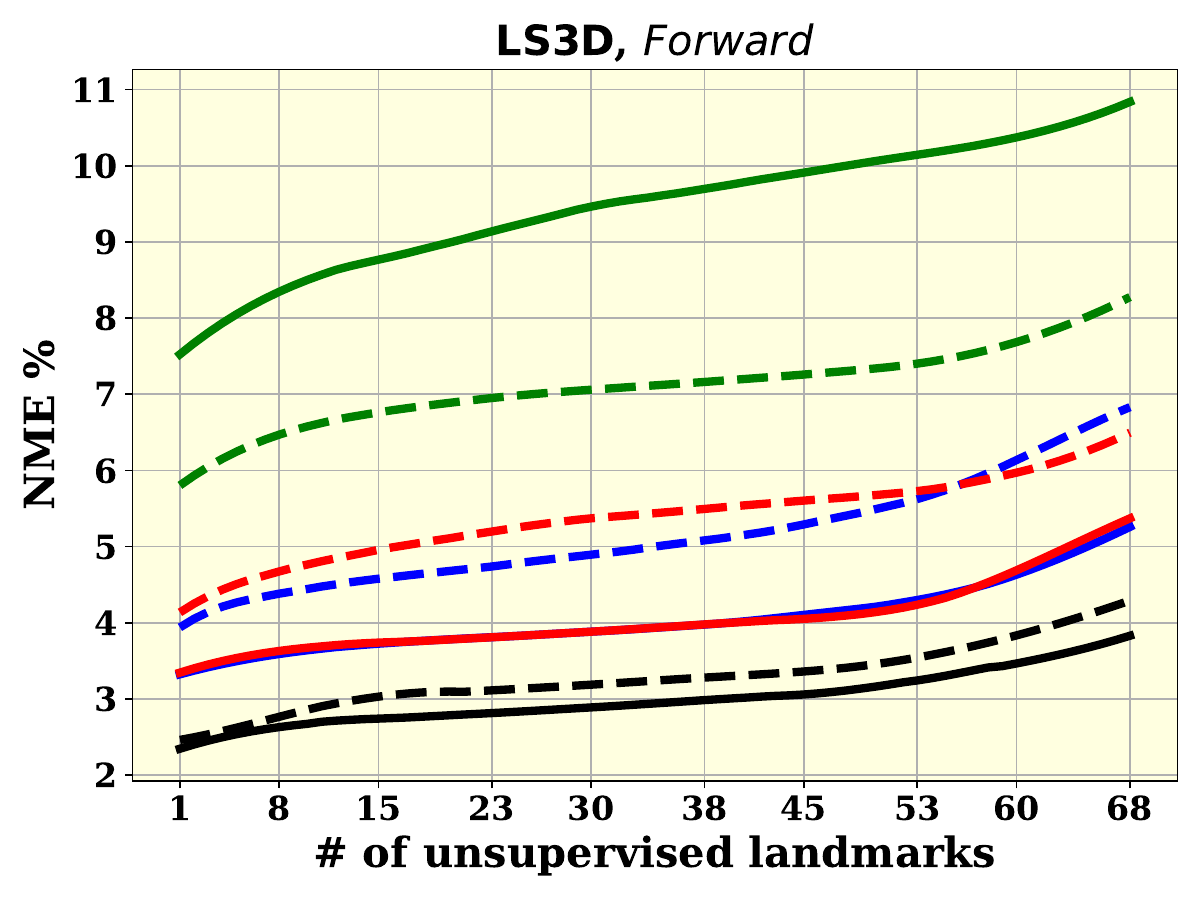}
    \label{fig:ls3d-fwd}
  }
  \subfloat{
    \includegraphics[width=0.21\textwidth]{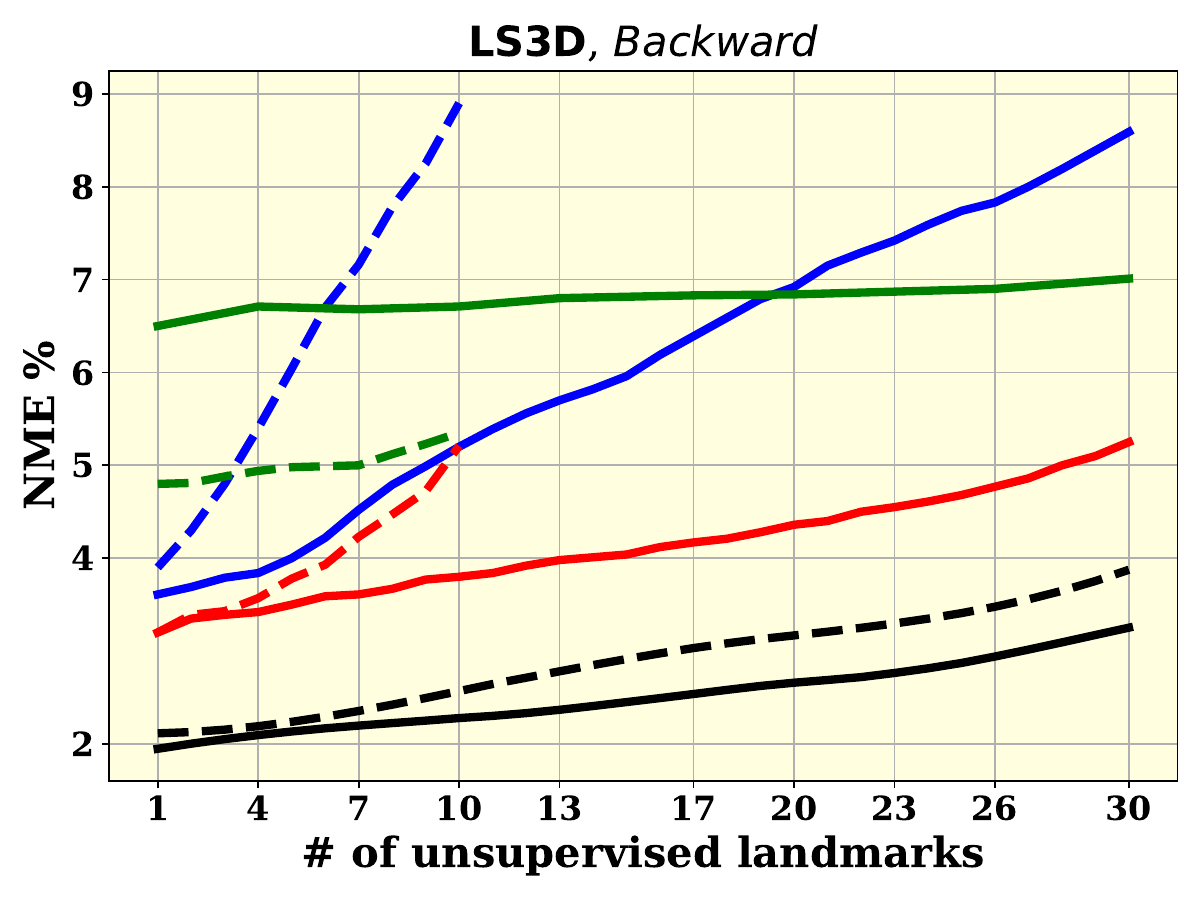}
    \label{fig:ls3d-fwd}
  }
  \caption{Cumulative Error Distribution (CED) Curves of forward and backward NME for MAFL and LS3D.}
  \label{fig:ced-curves} \vspace{-0.5em}
\end{figure*}

\noindent\textbf{Analysis:} 
As can be seen in~\cref{fig:method-quant} our D-ULD++ brings an improvement compared to D-ULD over all pose variations in aggregate. \cref{fig:method-qual} shows the variations and the learned landmarks are indeed better localized and more accurately matched across poses.

\section{Experiments}
\label{sec:experiments}

%---------------------------------------------------------------------------------------
%\subsection{Evaluation Metrics} % This will be changed, The original text will be written and replaced
%\noindent\textbf{Evaluation metrics \& baselines:}
\noindent\textbf{Datasets, Evaluation metrics \& Baselines:} We perform experiments on four diverse datasets: AFLW~\cite{koestinger2011annotated}, LS3D~\cite{bulat2017far}, CatHeads~\cite{zhang2008cat} and MAFL~\cite{liu2015deep}. We follow the same dataset splits and  evaluation protocol used in~\cite{mallis2023keypoints}. See supplementary material for details on datasets. 
For evaluation, we use the standard metrics of Forward and Backward Normalized Mean-squared Error NME\% widely used for ULD \cite{sanchez2019object, mallis2020unsupervised, mallis2023keypoints}. %This metric involves training a simple regressor, which maps the detected landmarks to ground-truth landmarks. 
In Forward NME, we train a regressor to map the discovered landmarks to the ground truth landmarks and compute NME\%. In the Backward NME, we train a regressor to map the ground truth landmarks to discovered landmarks and compute NME\%. %These two metrics collectively provide a comprehensive assessment of the accuracy and stability of our landmark detection method. 
The mappings are learned using a random subset of images from the dataset. We use the same subsets as~\cite{mallis2023keypoints}. 
%\noindent\textbf{Baselines:} 
We benchmark against the following baselines: Lorenz et al. ~\cite{lorenz2019unsupervised}, Shu et al.~\cite{sanchez2017functional}, Jakab et al.~\cite{jakab2018unsupervised}, Zhang et al.~\cite{zhang2018unsupervised}, Sanchez et al.~\cite{sanchez2019object}, Mallis et al.~\cite{mallis2023keypoints} and Awan et al.~\cite{awan2023unsupervised}. Where possible, we used pre-trained models, otherwise we re-trained these methods. Additionally, as all these methods do not make use of diffusion models, so we also retrain ~\cite{mallis2023keypoints} with Stable Diffusion features via the same aggregator network that we use instead of its FAN~\cite{bulat2017far} backbone. We refer to this method as Mallis (D). We use $K=10$ and $M=100$ as in the original method using the same training protocol.

\noindent\textbf{Training details:} Training is performed in the following sequence. \textbf{Bootstrapping: }The network $\Psi$ is initially trained with the Adam~\cite{kingma2014adam} optimizer with learning
rate $1\times 10^{-4}$
and betas $(0.9, 0.999)$; trained for $50K$ iterations, with a batch size of 12 per GPU. \textbf{D-ULD:} After bootstrapping, we alternate between clustering and training the network using pseudo-ground truth every $5000$ iterations. We train for a total of $100k$ iterations for all four datasets. The margin $m$ used in the contrastive loss is set to $0.8$. %We multiply the detector head loss with $0.2$ before training. 
\textbf{Pose-guided Proxy-Task:} For this stage, we initialize the model $\Psi$ with the training of D-ULD, append the VAE $\Psi_V$ to the descriptor head $\Psi_d$. For this stage, $\Psi_f$ is frozen and $\Psi_V$ is minimized by the ELBO loss. We reduce the learning rate for the Adam optimizer to $5 \times 10^{-5}$ and train for $50k$ iterations. \textbf{D-ULD++:} We discard the decoder of $\Psi_V$ and initialize the rest of the network weights with those obtained after proxy-task training. We train this network for $100k$ iterations using the Adam with the same parameters as before, but with a learning rate of $5\times 10^{-4}$. Every $5000$ iterations, we perform clustering and update the pseudo-ground truth. We report results for ZeroShot, D-ULD and D-ULD++ averaged over 5 evaluations and report relative gain throughout.

\begin{table*}[h]
\small
    \centering
    \begin{tabular}{cccccccccc}
   & \textbf{Method} & \multicolumn{2}{c}{\textbf{MAFL}} & \multicolumn{2}{c}{\textbf{AFLW}} & \multicolumn{2}{c}{\textbf{LS3D}} & \multicolumn{2}{c}{\textbf{CatHeads}}\\
      &    & F & B & F & B & F & B & F & B\\
                  \hline
      %Lorenz~\cite{lorenz2019unsupervised} (K=10) &  3.24 & \textbf{O} & \textbf{A} &  \textbf{B} & \textbf{C} & \textbf{D} &  9.30 & \textbf{F} \\
      %Shu~\cite{sanchez2017functional} & 5.45 & \textbf{O} &  \textbf{A} & \textbf{B} & \textbf{C} & \textbf{D} & \textbf{E} & \textbf{F}\\
    \multirow{5}{*}{\rotatebox[origin=c]{90}{Published}} & Jakab~\cite{jakab2018unsupervised} (K=10) &  \underline{3.19} & 4.53 & 6.86 & 8.84 & 5.38 & 7.06 & 4.53 & 4.06 \\
     & Zhang~\cite{zhang2018unsupervised} (K=10) & 3.46 & 4.91 & 7.01 & 8.14 & 6.74 & 7.21 & 4.62 &  4.15\\
     & Sanchez~\cite{sanchez2019object} (K=10) & 3.99 & 14.74 &  6.69 & 25.84  & 26.41 & 5.44 & 4.42 & 4.17\\
%      Sahasrabudhe~\cite{sahasrabudhe2019lifting} & 6.01 & \textbf{O} & \textbf{E} & \textbf{F} & \textbf{A} & \textbf{B} & \textbf{C} & \textbf{D} \\
     & Mallis~\cite{mallis2023keypoints} (K=10)&  \underline{3.19}   & \underline{4.23} & 7.37 & 8.89 &  6.53 & 6.57 &  9.31 & 10.08 \\
    &  Awan~\cite{mallis2023keypoints} &  3.50   & 5.18 & \underline{5.91} &  \underline{7.96} & \underline{5.21} & \underline{4.69} & \underline{3.76} & \underline{3.94} \\
      \hline
    &  Mallis (D) (K=10)& 2.74 & 3.11 &  3.38 & 3.75 & 2.89 &  3.76 & 3.14 & 3.62\\
    &     Zero Shot (K=10) & 3.14 & 3.27  &  4.98   & 6.29  & 3.53 &  4.14 & 3.47 & 3.59\\
      &    &{\scriptsize \color{blue} +1.19\%} & {\scriptsize \color{blue} +6.14\%} &  {\scriptsize \color{blue} +15.74\% }& {\scriptsize \color{blue} +20.97\%} & {\scriptsize \color{blue} +34.35\%} & {\scriptsize \color{blue} +11.72\%} & {\scriptsize \color{blue} +7.7\%} & {\scriptsize \color{blue} +8.88\%}\\
    &    D-ULD++ (K=10) & \textbf{2.19} & \textbf{2.78} &  \textbf{2.92} & \textbf{3.62} & \textbf{2.12} & \textbf{2.85} & \textbf{2.89} & \textbf{3.12}\\
     &   & {\scriptsize \color{blue} +31.34\%} & {\scriptsize \color{blue} +34.28\%} & {\scriptsize \color{blue} +50.91\%} & {\scriptsize \color{blue} +54.52\%} & {\scriptsize \color{blue} +59.31\%} & {\scriptsize \color{blue} +39.23\%} & {\scriptsize \color{blue} +23.13\%} & {\scriptsize \color{blue} +20.81\%}\\
    \end{tabular}
    \caption{Error comparison on MAFL, AFLW, LS3D and CatHeads, in Forward and Backward NME (denoted as F and B). The
results of other methods are taken directly from the papers (for the case where all MAFL training images are used to train
the regressor and the error is measured w.r.t. to 5 annotated points).  A set of 300 training images is used to train the regressors.
Error is measured w.r.t. the 68-landmark configuration typically used in face alignment. The best performance is in \textbf{bold}.  Below ZeroShot and D-ULD++ are the (relative) percentage improvements shown in blue over the best published results (shown underlined).}
    \label{tab:results}
    \vspace{-2em}
\end{table*}

\begin{table}[h]
\small
    \centering
    \begin{tabular}{ccccc}
    \textbf{Method} & \multicolumn{2}{c}{\textbf{AFLW}} & \multicolumn{2}{c}{\textbf{MAFL}}\\
          & F & B & F & B\\
                  \hline
         ZeroShot & 
         4.98 & 6.29 & 3.14 & 3.98\\
         D-ULD & 3.42 & 4.89 & 2.82  & 3.21  \\
         + PP  &  3.15 & 4.46 & 2.68 & 3.12\\
         + (TSC w/ Full VAE) & 3.07 & 4.12 & 2.59 & 3.03 \\
         + TSC (D-ULD++) &  \textbf{2.92} & \textbf{3.62} & \textbf{2.19} & \textbf{2.78}\\
    \end{tabular}
    \caption{Ablation Evaluation on AFLW and MAFL. PP stands for pose-guided proxy task and TSC stands for two-stage clustering. }%F and B stand for forward and backward NME\%.}
    \label{tab:ablation}
    \vspace{-0.7em}
\end{table}

\begin{table}[h]
    \scriptsize
    \centering
    \begin{tabular}{ccccc}
       \multirow{2}{*}{\textbf{Methods}}  & \multicolumn{2}{c}{\textbf{MAFL}} & \multicolumn{2}{c}{\textbf{AFLW}} \\
         & Sil. & CH & Sil. & CH\\
         \hline 
      ZeroShot ($K=30$)  & 0.74 & 357.4 & 0.72  & 142.9\\
      D-ULD ($K10$)  & 0.79 & 372.1 & 0.81  & 182.3\\
      D-ULD ($K=30$)   & 0.82 & 382.6 & 0.83  & 190.7\\
      D-ULD++ ($Q=10,K=10$)  & 0.86 & 377.4 & 0.86  & 192.3\\
      D-ULD++ ($Q=10,K=30$)   & 0.87 & 392.6 & 086  & 194.7\\
    \end{tabular}
    \caption{Quality of clustered landmark representations %for ZeroShot, D-ULD and D-ULD++ 
    using Silhouette coefficient (Sil.) and Calinski-Harabasz (CH) Index. $Q$ and $K$ are the number of pose 
 and  keypoint clusters.}
    \label{tab:clustering-coefficients} 
    \vspace{-0.7em}
\end{table}

% \begin{figure}[h]
%     \centering
%   \subfloat[ ]{\includegraphics[width=0.21\textwidth]{./figs/mafl_num_clusters.pdf}
%     \label{fig:mafl-num-clusters}}
%   \subfloat[]{
%    \includegraphics[width=0.21\textwidth]{./figs/aflw_num_clusters.pdf}
%     \label{fig:aflw-num-clusters}
%   }
%   \caption{Impact of number of clusters on performance. Along the X-axis and Y-axis number of clusters for keypoints and poses are shown respectively. For each permutation of number of clusters the forward and backward NME is shown as a tuple.}
%   \label{fig:cluster-impact}
% \end{figure}

\subsection{Results}
\Cref{tab:results} summarizes the results of our approach and baselines. The simple ZeroShot baseline outperforms previously published methods across all datasets, notably surpassing them by over $30\%$ on LS3D and $10\%$ on AFLW for both forward and backward NME respectively.  This emphasizes the efficacy of Stable Diffusion features.
%, which is unsurprising given that earlier methods depended on less expressive backbone networks. Further, the results of ZeroShot argue for a re-orientation of unsupervised landmark detection and related tasks from more complex architectures and training schemes towards exploring more effective ways to utilize existing the feature representations of pre-trained large models.
%
The interesting comparison is between Mallis (D) and D-ULD++. 
%as both methods have common network features  and a two-stage self-training based supervision, with similar intitial stages. 
%The key difference in both methods is the second stage of self-training based supervision. 
Mallis (D) simply oversegments the initial clusters into more fine-grained clusters (the initial $K$ clusters are segmented into $M>>K$ clusters), while D-ULD++ proposes a pose-guided proxy task and a two-stage clustering.
D-ULD++ outperforms Mallis (D) by notable margins. It also shows a relative improvement of more than $20\%$ over published methods on all four datasets. %This performance is achieved without any fine-tuning of the Stable Diffusion network. 

%As can be clearly seen, D-ULD++ significantly outperforms the other methods margins of more than $50\%$ on LS3D, MAFL and AFLW. It also performs $30\%$ better than competing methods on CatHeads. The main reason for this is, the features we use are trained on much greater quantities of data and are thus more expressive.

%We quantified the impact of these features using a zero-shot model (denoted ZeroShot in~\Cref{tab:results}) by randomly selecting keypoints from dataset images, extracting features from a pre-trained Stable Diffusion model, and clustering them with $K$-Means. For each image, we retained the closest descriptor to each cluster centroid along with its associated keypoint. As can be clearly seen, even a simple Zero-shot baseline improves upon previous results by a margin of above $X\%$ for all datasets. Adding our proposed self-training scheme (Ours in~\Cref{tab:results}) only further improves upon previous methods. We believe this sets a new paradigm for unsupervised landmark detection as the features from these networks achieve results far exceeding previous methods.  This highlights the emergent properties of training large models on vast amounts of data for image generation can also be utilized for tasks like unsupervised landmark detection with simple training schemes.
\begin{figure}[!htp]
    \centering
\includegraphics[height=0.35\textwidth]{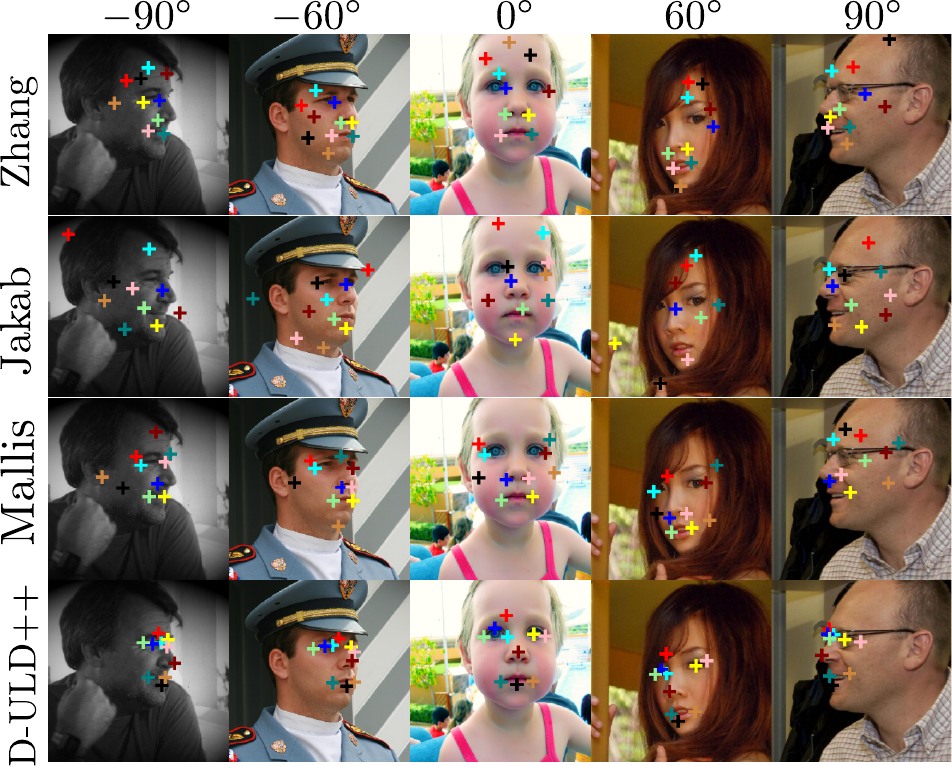}
    \caption{Landmarks discovered across poses in LS3D.}
    \label{fig:qual-results} \vspace{-2em}
\end{figure}
\Cref{fig:ced-curves}  displays the Cumulative Error Distribution (CED) curves. %comparing various methods to D-ULD++. 
Our D-ULD++ exhibits notably lower errors than others. Compared to others, our method's curve displays a gradual slope, suggesting a gradual decline in keypoint localization accuracy. In contrast, the curves for other methods start with higher base errors and exhibit steeper slopes, indicating a failure in keypoint detection/localization accuracy beyond a certain number of keypoints.
\Cref{fig:qual-results} provides a visual comparison of landmarks discovered by D-ULD++ and various baselines on images across an almost $180^\circ$ pose range. % Jakab and Zhang completely fail to model large viewpoint changes learning mostly random landmarks in uniform areas and not repeatable edges. Mallis manages to accurately detect a few keypoints across the poses like the lip edges,  but fails on other keypoints. This is probably due a combination of the poorer Stage2 self-training than D-ULD++, as well as less expressive features. D-ULD++ is to reliably keypoints across $180^\circ$ pose variations.
\Cref{fig:hungarian-plot} gives further insight into the localization accuracy and reliability of the landmarks detected by D-ULD++. In MAFL, which has for each image 68 facial landmarks, each landmark is aligned with the best-matching unsupervised landmarks using the Hungarian algorithm. A value of $K=30$ is employed across all methods. Notably, our detected landmarks exhibit top accuracy in tracking semantically relevant facial landmarks. D-ULD++ has almost perfect accuracy in detecting keypoints in the eyebrow, nose, eyes and mouth region.

%To further demonstrate that, we evaluate how accurately raw unsupervised landmarks track supervised landmark locations in Fig. 8.

% \subsection{Pose Analysis}
% We use a trained pose regressor to quantify whether 

\section{Ablation and Analysis}
\label{sec:ablation}

\noindent\textbf{Performance contribution of our methods:} Over ZeroShot baseline, our D-ULD provides consistent improvement in both forward and backward NME (\Cref{tab:ablation}). Upon introducing the pose-guided proxy task (PP) in D-ULD, we see a further improvement in all instances. After including our two-stage clustering (TSC) with PP in D-ULD, we note a consistent notable improvement over each of the variants. To assess the contribution of~\Cref{eqn:latent-opt} we compare it to training the network with the full VAE. The VAE output is supervised by the ELBO loss. D-ULD++ is superior to full VAE alternative, justifying use of~\Cref{eqn:latent-opt}. %over the commonly used ELBO loss.

\begin{figure}[h]
    \centering
\includegraphics[scale=0.30]{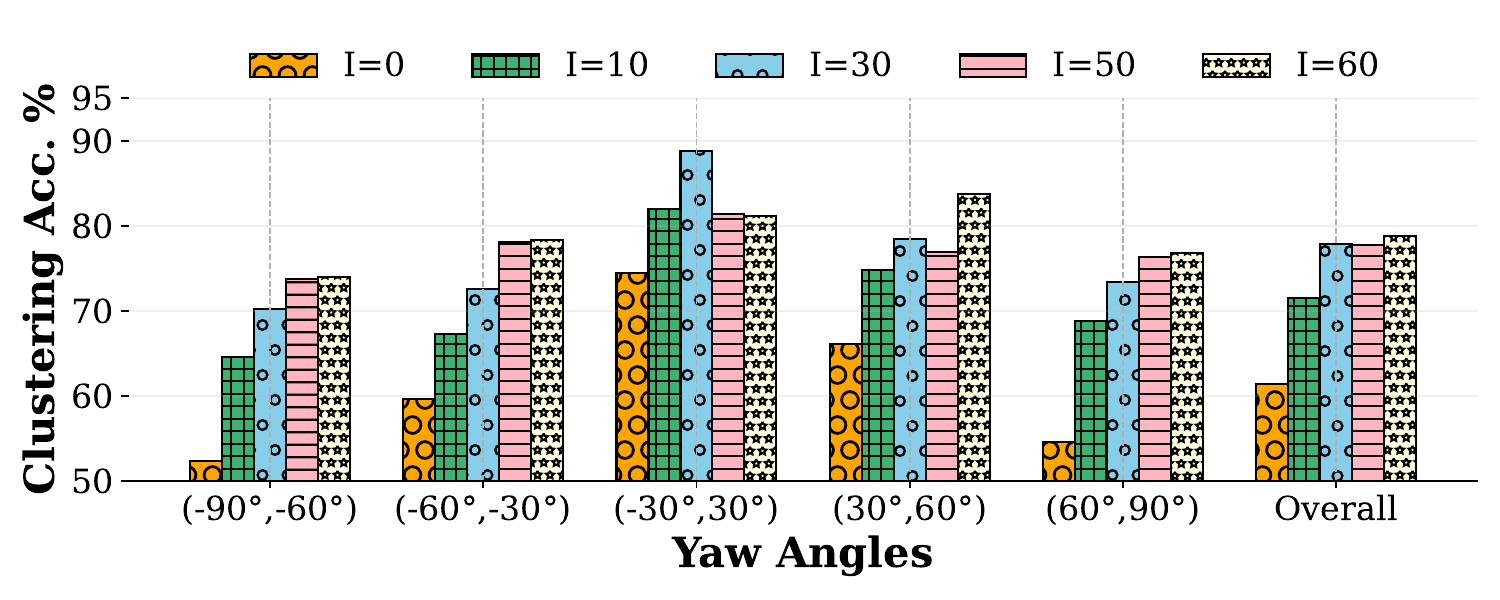}
    \caption{Clustering accuracy percentages are reported across different yaw-angle ranges and overall for D-ULD++ on LS3D. The variable $I$ denotes the iteration number, as a multiple of 1000.}
    \label{fig:cluster-analysis} \vspace{-1em}
\end{figure}

\noindent\textbf{Effect of TSC-guided Self-Training:} 
To get a better understanding of whether the pseudo-labels from TSC do indeed capture pose variations, we do a simple cluster analysis. Recall D-ULD++ generates a latent pose code for each image landmarks, \ie there is a \emph{1-to-1} relation between the latent code and image landmarks. These latent codes are then clustered into $Q$ clusters. As the LS3D images come with the 5 yaw-angle ranges (the ranges are shown in~\Cref{fig:cluster-analysis}) they belong to, we cluster the latent codes into $Q=5$ clusters. For the latent codes within each cluster, we find which yaw-angle range the  majority of the latent codes belong to (by checking the range of their corresponding images) and map each cluster to a yaw-angle range. We find that the $Q=5$ clusters neatly map to each of the $5$ ranges. In the process, we also measure the percentage of latent codes that lie within the clusters assigned range. We refer to this percentage as clustering accuracy $\%$. \Cref{fig:cluster-analysis} shows the cluster accuracy for $K=10$ clusters as a function of iterations for the various pose ranges. As can be clearly seen, clustering accuracy increases up to iteration $50k$ before levelling off.
%
%~\Cref{tab:ablation} compares our two-stage clustering (TSC) in D-ULD++ with simple clustering in D-ULD. We see that, our TSC mostly improves over simple clustering variants in both clustering quality metrics \ie Silhouette coefficient and Calinski-Harabasz (CH) Index indicating more compact and internally consistent clusters..
%
\Cref{tab:ablation} shows TSC outperforms the  simple clusterinig in D-ULD, achieving a higher Silhouette coefficient and Calinski-Harabasz (CH) Index. This indicates TSC's superiority in forming compact clusters.
%and internally consistent clusters.

%maximum number of elements within the cluster fall into. This way we map each cluster to a single yaw-angle range. We compute 

%$$ shows the percentage of the images within a cluster that lie within the same yaw ranges for LS3D. We show the yaw-angle range-wise split as well as overall accuracy. As the value of $Q$ is increased from $10$ to $30$, the overall accuracy $\%$ increases $XX\%$. Also, we show the accuracy $\%$ after various epochs. Pose pseudo-Label self-training improves the overall accuracy by $\%$ at epoch 30 with significant improvements for the more challenging side-profile ranges of $XX\%$ $XX\%$ , \ie $(-90^\circ,-60^\circ)$  and  $(60^\circ,90^\circ)$ respectively. 

\begin{figure}[!htp]
    \centering
  \subfloat[ ]{\includegraphics[width=0.22\textwidth]{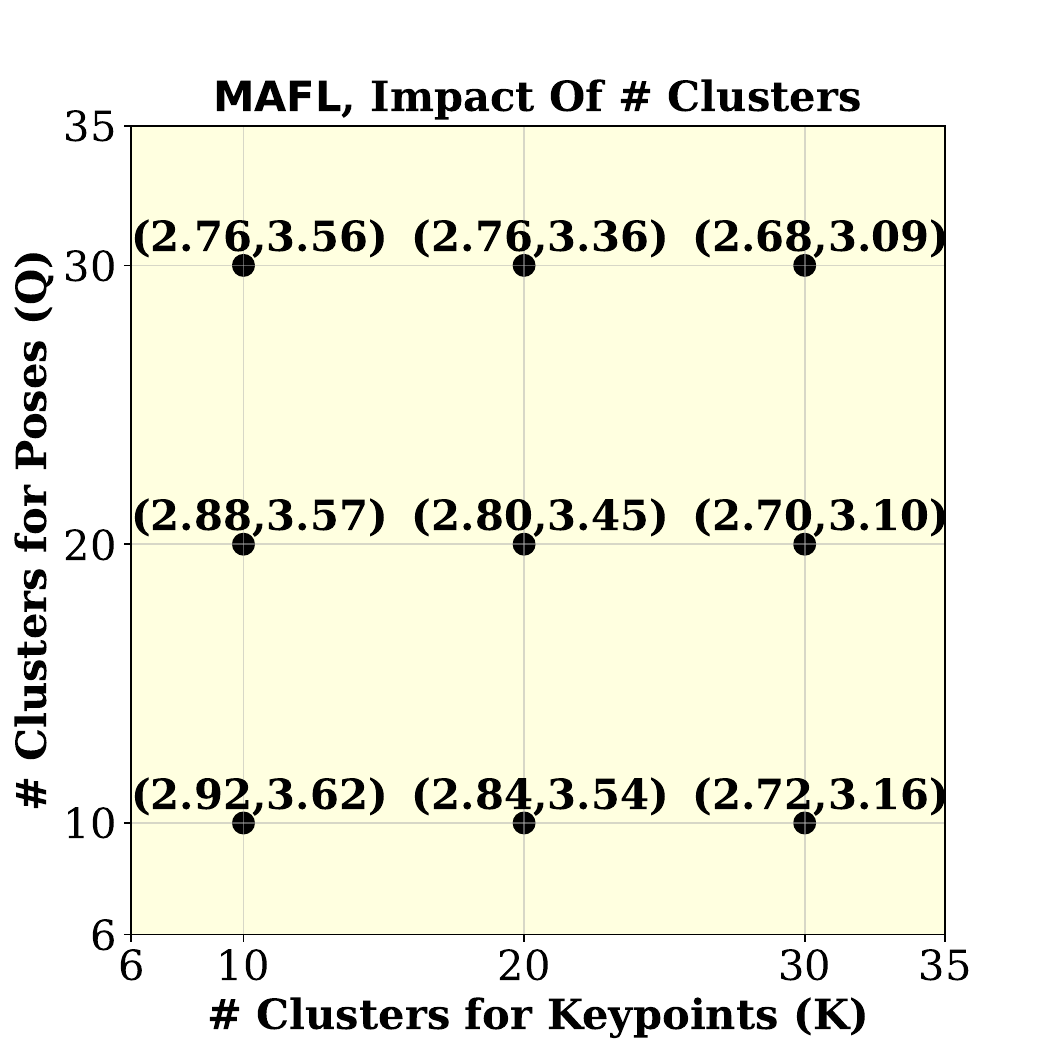}
    \label{fig:mafl-num-clusters}}
  \subfloat[]{
   \includegraphics[width=0.22\textwidth]{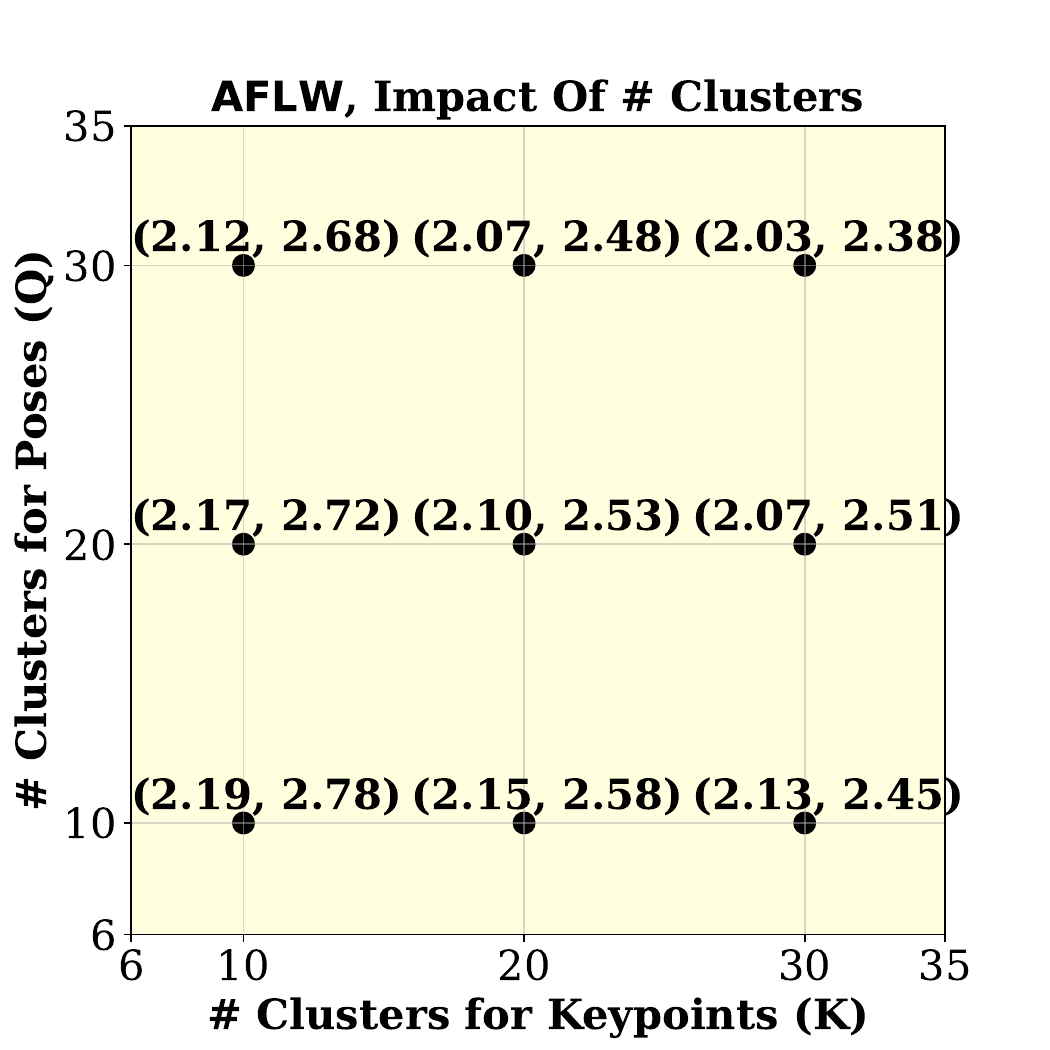}
    \label{fig:aflw-num-clusters}
  }
  \caption{Impact of number of clusters on performance. Along the X-axis and Y-axis number of clusters for keypoints and poses are shown respectively. For each permutation of number of clusters the forward and backward NME is shown as a tuple.}
  \label{fig:cluster-impact} 
  \vspace{-0.7em}
\end{figure}

\noindent\textbf{Hyper-Parameter Study:}~\Cref{fig:cluster-impact} shows the effect of varying the number of clusters for k-Means on Forward and Backward NME for both keypoint and pseudo-pose clustering. $K$ and $Q$ are the  number of clusters for keypoint and pseudo-pose clustering respectively (\Cref{sec:method}). 
Increasing $Q$ and $K$ has the effect of lowering both NMEs. As $K$ increases the keypoints captured with each cluster are more specific and localized. Similarly, increasing $Q$ would have the effect of capturing finer pose variations. This saturates at $Q=30$ and $K=30$, but this maybe due to the limitation of the dataset size.

\section{Conclusion}
\label{sec:conclusion}
We explored the effectiveness of Stable Diffusion to tackle unsupervised landmark detection (ULD) problem. We first proposed a ZeroShot baseline that is only based on clustering of diffusion features and nearest neighbour querying that excels in performance than SOTA methods. 
This motivated us to develop ULD algorithms that involves fine-tuning diffusion features. We propose D-ULD which shows the effectiveness of fine-tuning and D-ULD++ which proposes a novel proxy task and two-stage clustering mechanism based on this proxy task. Our methods consistently achieves significant improvement over existing baselines.

%\paragraph{Acknowledgements}Dr. Arif Mahmood is funded by Information Technology University of the Punjab.

% Leveraging the remarkable power of diffusion models, we propose a novel self-training based scheme for unsupervised discovery of landmark objects. 

% We presented a novel path for unsupervised discovery ofobject landmarks based on two ideas, namely self-training and recovering correspondence. The former helps our system improve by using its own predictions and constitutes a natural fit for training an object landmark detector starting from generic, noisy keypoints. The latter, although being a key property of object landmarks detectors, has not been previously used for unsupervised object landmark discovery. Compared to previous works, our approach can learn view-based landmarks that are more flexible in terms of changes in 3D viewpoint, providing superior results on a variety of challenging facial and human pose datasets.

{
    \small
    \bibliographystyle{ieeenat_fullname}
    \bibliography{main}
}

%%%%%%%%%%%% Supplementary Material Start %%%%%%%%%%%%%%
\clearpage
\setcounter{page}{1}
\maketitlesupplementary
\section{Pseudo-code D-ULD++}
The architecture for D-ULD++ is shown in Fig.~2 main manuscript. %~\Cref{fig:duldpp-architecture}. 
The input to the architecture is an image $\mathbf{x}_j$. The aggregator network $\Psi_b$ branches into the descriptor head  and the detector head with the VAE auto-encoder appended to it. The output of the descriptor head is given by the following sequence of operations $\mathbf{F}^j = \Psi_f(\Psi_b(\mathbf{x}_j))$. The operations for the modified detector head is given by $\boldsymbol{l}_j = \Psi_V^{Enc}(\Psi_d(\Psi_b(\mathbf{x}_j)))$.

The following contrastive loss is minimized for the descriptor head.

\begin{align}
\SL_{\mathbf{f}}(\mathbf{f}^j_i,\mathbf{f}^{j'}_{i'}) = \mathbf{1}_{[\mathbf{c}^j_i = \mathbf{c}^{j'}_{i'}]} ||\mathbf{f}^j_i - \mathbf{f}^{j'}_{i'}||\text{ + }\nonumber \\ \mathbf{1}_{[\mathbf{c}^j_i \neq \mathbf{c}^{j''}_{i''}]} \text{max}(0,m-||\mathbf{f}^j_i-\mathbf{f}^{j''}_{i''}||)
\label{eqn:descriptor-loss}
\end{align}
Descriptors with the same labels $\mathbf{c}^j_i = \mathbf{c}^{j'}_{i'}$ are pushed together, whereas those with different are minimized unless separated by a margin $m$.

Likewise for the detector head, we minimize the following loss: 
\begin{align}
\SL_{\boldsymbol{\varphi}}(\boldsymbol{\varphi}_j,\boldsymbol{\varphi}_{j'}) = \mathbf{1}_{[\boldsymbol{u}_j = \boldsymbol{u}_{j'}]} ||\boldsymbol{\varphi}_j - \boldsymbol{\varphi}_{j'} || + \nonumber \\ \mathbf{1}_{[\boldsymbol{u}_j \neq \boldsymbol{u}_{j''}]}\text{max}(0,m-||\boldsymbol{\varphi}_j-\boldsymbol{\varphi}_{j''}||)
\label{eqn:latent-loss}
\end{align}

\Cref{eqn:latent-loss} pushes latent codes with the same labels together, \ie $\boldsymbol{u}_j = \boldsymbol{u}_{j'}$. 

The pseudo-code for D-ULD++ is described in~\Cref{alg:pseudo-code}.

\label{sec:pseudo-code}
\begin{algorithm*}
\caption{Update-Dataset $\SX$}\label{alg:update-dataset}
\begin{algorithmic}
\Require $\SX = \{\mathbf{x}_j \mid j \in \text{images} \}$
\State 1. $\{\mathbf{p}^j_i,\mathbf{f}^j_i\}_{i \in N_j} = \text{Extract keypoints and descriptors from}~\Psi(\mathbf{x}_j)$ \Comment{\small Keypoints and descriptors are extracted for each image $\mathbf{x}_j$.}
\State 2. $\SX=\{\mathbf{x}_j,\{\mathbf{p}^j_i,\mathbf{f}^j_i,\mathbf{c}^j_i \}_{i=1}^N \}$~\Comment{Update $\SX$ with keypoints, descriptors and cluster pseudo-labels.$\{\mathbf{p}^j_i,\mathbf{f}^j_i,\mathbf{c}^j_i\}_{i \in N_j}$.}
\State 3. $\boldsymbol{\varphi}_j=\Psi_V^{Enc}(\Psi_d(\Psi_b(\mathbf{x}_j)))$~\Comment{Extract the latent codes for each image  $\mathbf{x}_j$.}
\State 4. $\boldsymbol{l}_j$ = KMeans($\{\boldsymbol{\varphi}_j\}$)~\Comment{Compute pose latent-code cluster labels $\boldsymbol{l}_j$.}
\Ensure $\SX = \{\mathbf{x}_j,\{\mathbf{p}^j_i, \mathbf{f}^j_i, \mathbf{c}^j_i\}, \boldsymbol{l}_j, \boldsymbol{u}_j\}.$
\end{algorithmic}
\end{algorithm*}

\begin{algorithm*}
\caption{Pseudo-Code D-ULD++}\label{alg:pseudo-code}
\begin{algorithmic}
\State $\SX=$Update-Dataset($\SX$) \Comment{$\SX$ is updated. $\SX = \{\mathbf{x}_j,\{\mathbf{p}^j_i, \mathbf{f}^j_i, \mathbf{c}^j_i\}, \boldsymbol{l}_j, \boldsymbol{u}_j\}.$}

\State{\textbf{Main Training Loop}}
\For {epoch = $1 \rightarrow N_E$}~\Comment{Epoch loop.}
\For{$i = 1 \rightarrow N_{it}$}~\Comment{Iterate for $N_{it}$ iterations.} 
\State $\{ \mathbf{x}_j, \{ \mathbf{p}^j_i,\mathbf{f}^j_i,\mathbf{c}^j_i \}, \boldsymbol{l}_j, \boldsymbol{\varphi}_j \}$ = GetBatch($\mathbf{x}_j$)
\State Update the network $\Psi$, $\Psi_V^{Enc}$ with the gradients of$\SL_f$ and $\SL_{\boldsymbol{\varphi}}$.

% Compute losses and $\nabla$ for $\SL_f$ (~\Cref{eqn:descriptor-loss}) and $\SL_\phi$ (~\Cref{eqn:latent-loss}) and update network.
% \State 
\EndFor
\State 5. Re-populate $\SX$ by redoing steps $1$ to $4$. $\SX = \{\mathbf{x}_j,\{\mathbf{p}^j_i, \mathbf{f}^j_i, \mathbf{c}^j_i\}, \boldsymbol{l}_j, \boldsymbol{u}_j\}$
\EndFor
\end{algorithmic}
\end{algorithm*}

\section{Consistency Analysis}
\label{sec:consistency}

We perform consistency analysis to evaluate whether the detected landmarks are consistent or not ~\cite{sanchez2019object}. The consistency of detected landmarks is defined as, $e_k = || \Psi_d(\Psi_b(A(\mathbf{x}_j)))-A(\Psi_d(\Psi_b(\mathbf{x}_j)))||)$, where $A$ is a random similarity transformation. $\Psi_d$ and $\Psi_b$ are the descriptor head and aggregator network respectively.

%$pi_k$ denotes the $k^{th}$ land-ark extracted by the network.
%
% To evaluate whether or not the landmarks detected are stable, we perfor- a stability analysis. The stability of discovered land-arks~\cite{sanchez2019object} as, $e_k = || \pi_k(A(y))-A(\pi_k||y||)$, where $A$ denotes a rando- si-ilarity transformation. $pi_k$ denotes the $k^{th}$ landmark extracted by the network.
% %
We report consistency errors, averaged over $K=10$ landmarks, in~\Cref{tab:stability}. Our method produces more consistent landmarks than the competing approaches on all datasets.

\begin{table}[!htp]
    \centering
    \begin{tabular}{ccccc}
        \textbf{Method} & \textbf{MAFL} & \textbf{AFLW} & \textbf{CatHeads} & \textbf{LS3D} \\
        \hline
        Sanchez~\cite{sanchez2019object} &  8.78 & 7.56 & 2.58 & 21.3\\
        Awan~\cite{awan2023unsupervised} & 2.37 & 1.77 & 2.24 & 3.23\\
        D-ULD++ (Ours) & 1.56 & 0.87 & 1.78 & 1.98\\
    \end{tabular}
     \caption{Our method (D-ULD++) produces more consistent landmarks than the competing methods across all datasets.}
    \label{tab:stability}
\end{table}

\section{Additional CED Curves}
\label{sec:ced-curves}
\begin{figure}[!htp]
    \centering
  \subfloat{
    \includegraphics[width=0.4\textwidth]{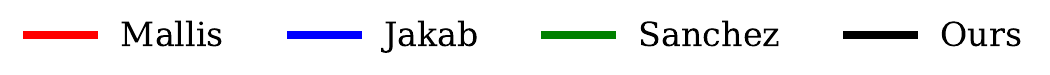}
    \label{fig:legend-ced}}\newline
  \subfloat{
    \includegraphics[width=0.23\textwidth]{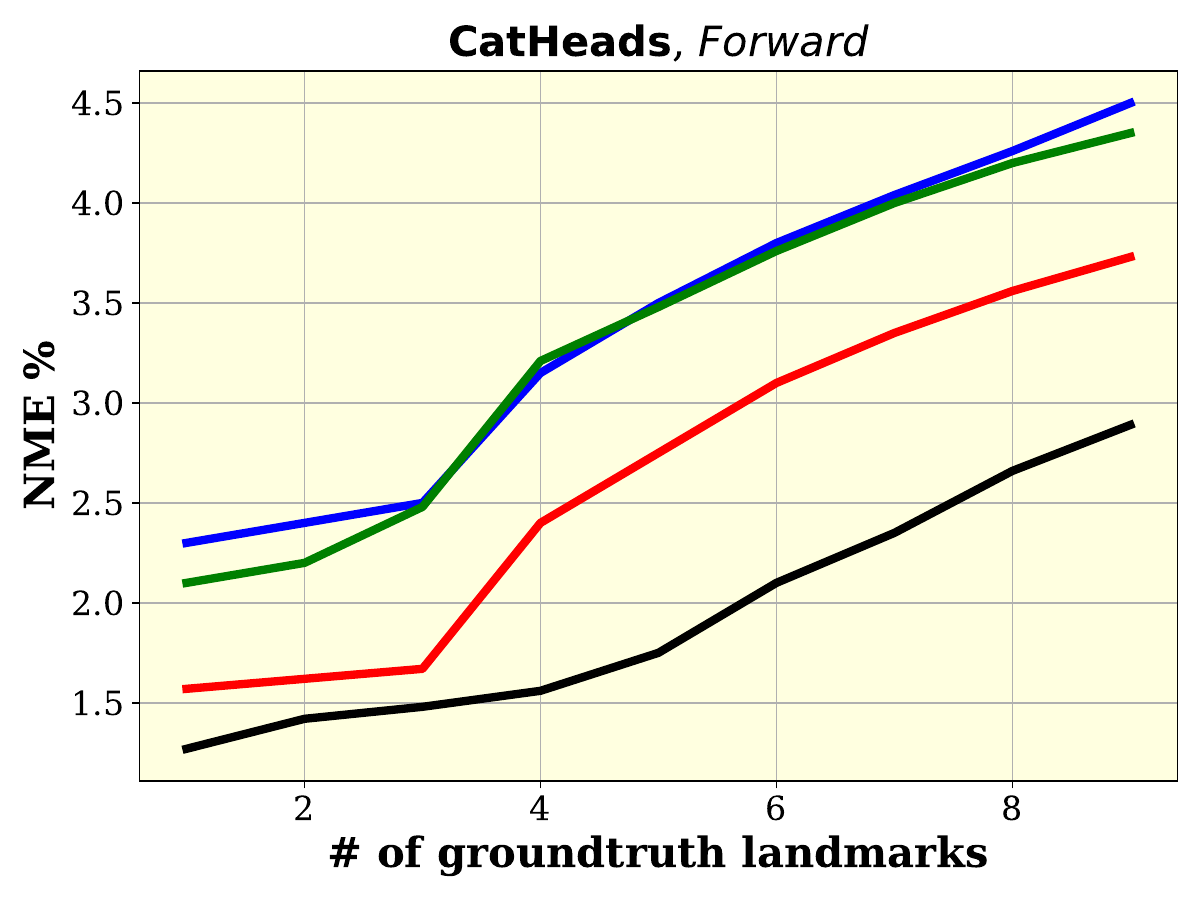}
    \label{fig:mafl-fwd}}
  \subfloat{
    \includegraphics[width=0.23\textwidth]{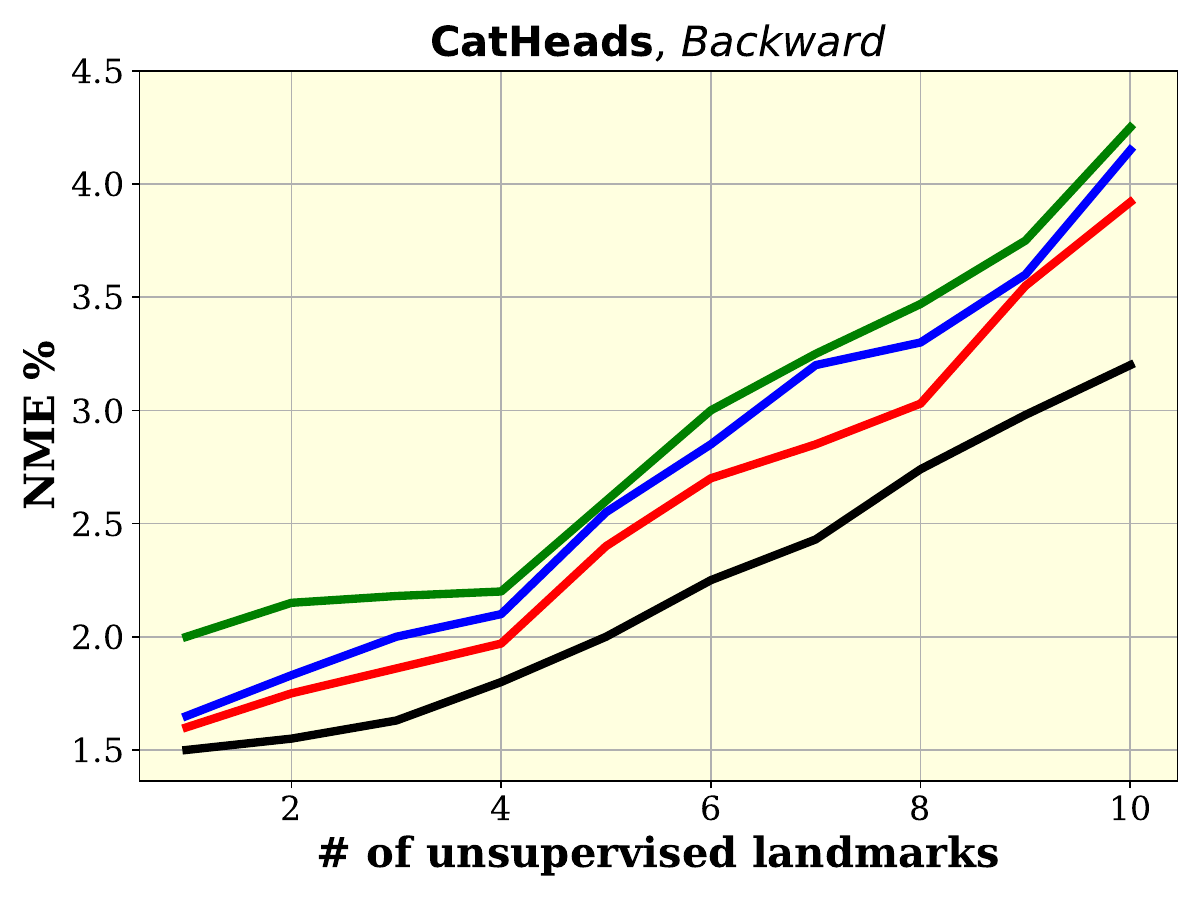}
    \label{fig:mafl-bwd}
  }\newline
  \subfloat{
    \includegraphics[width=0.23\textwidth]{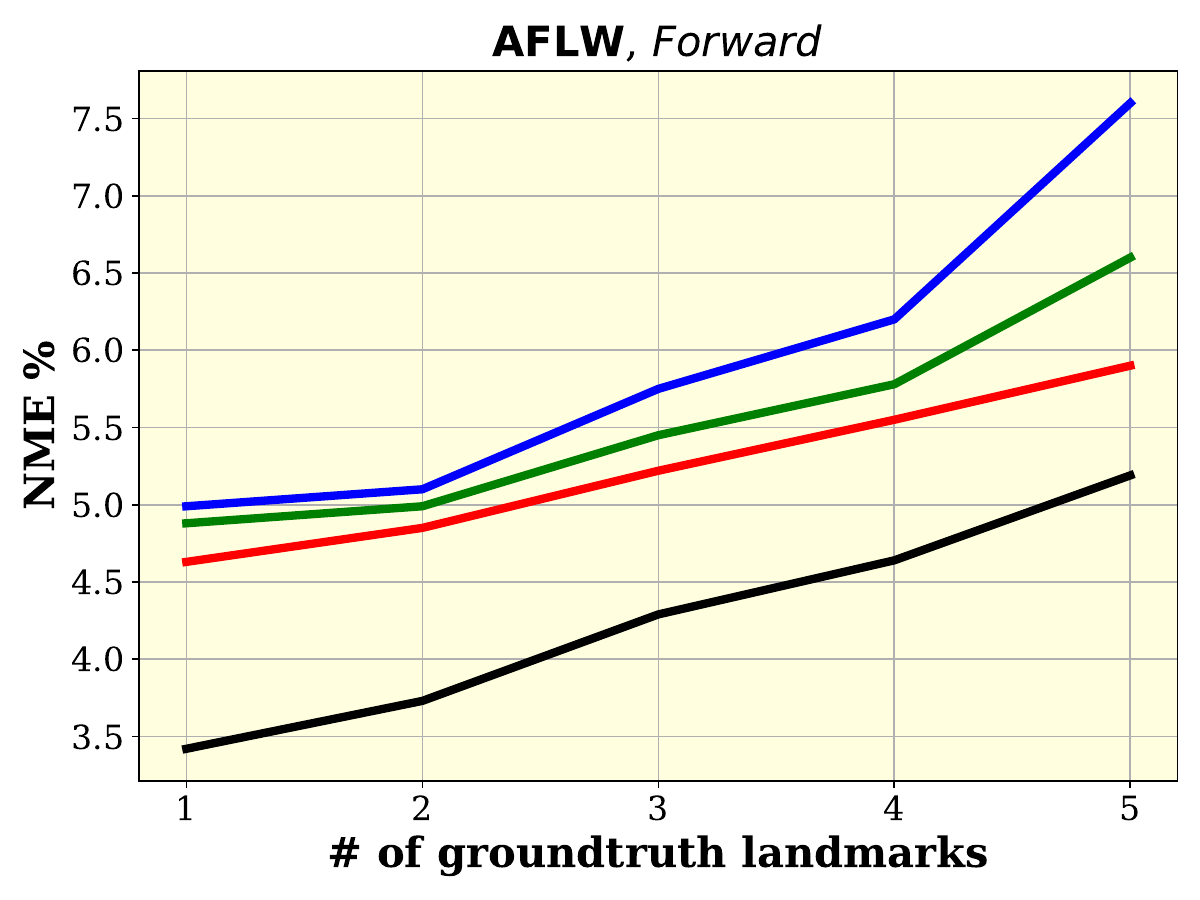}
    \label{fig:ls3d-fwd}
  }
  \subfloat{
    \includegraphics[width=0.23\textwidth]{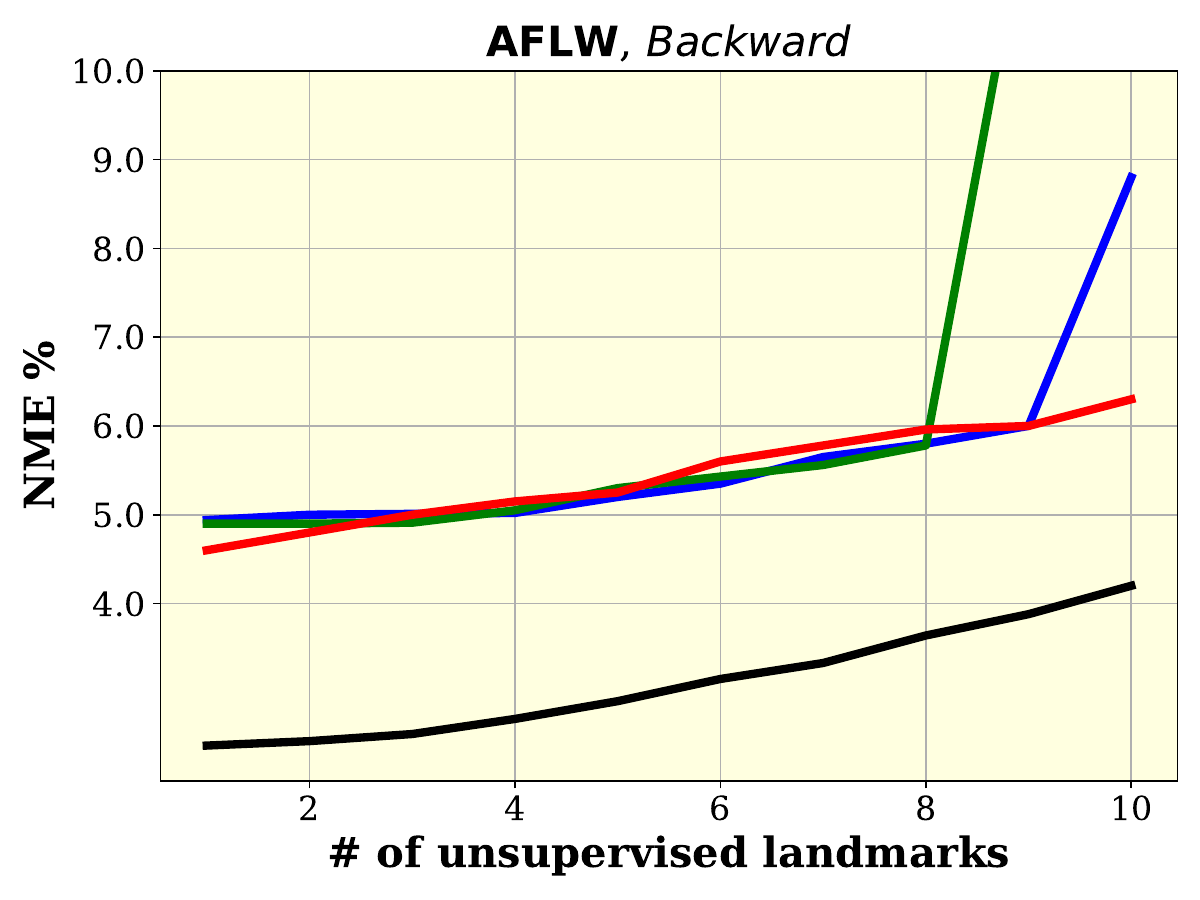}
    \label{fig:ls3d-fwd}
  }
  \caption{Cumulative Error Distribution (CED) Curves of forward and backward NME for CatHeads and AFLW.}
  \label{fig:additional-ced-curves} \vspace{-0.5em}
\end{figure}

\Cref{fig:additional-ced-curves} shows the cumulative error curves (CED) curves for CatHeads and AFLW datasets. In concurrence with the CED curves from the main manuscript, our method shows significantly lower base error and a more gradual degredation in performance.

\section{Qualitative Results}
\label{sec:qual-results}
\begin{figure*}
    \centering
    \includegraphics[scale=0.9]{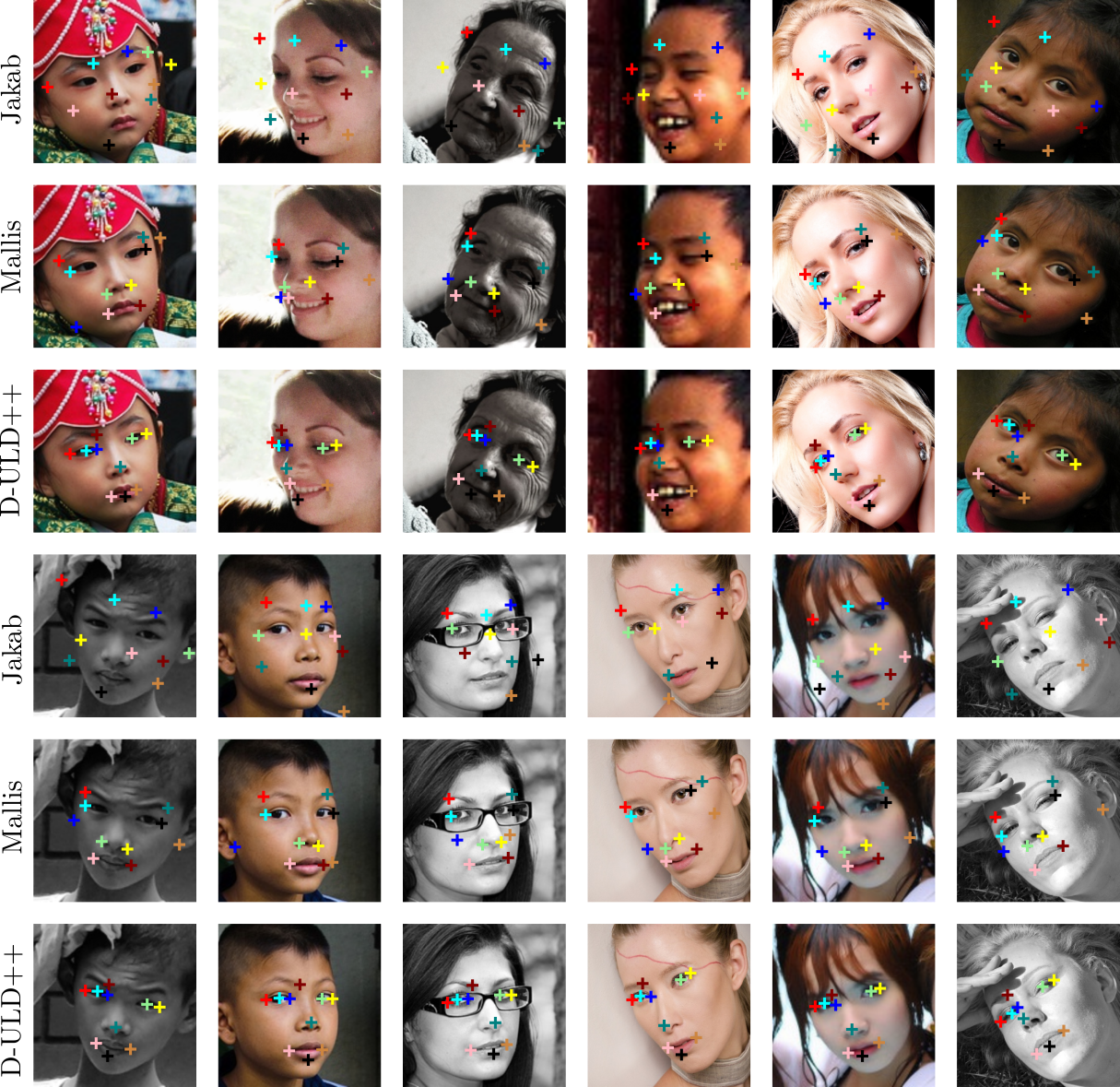}
    \caption{Results Comparison on AFLW for Jakab, Mallis and D-ULD++.}
    \label{fig:aflw-results-supp}
\end{figure*}

\begin{figure*}
    \centering
    \includegraphics[scale=0.9]{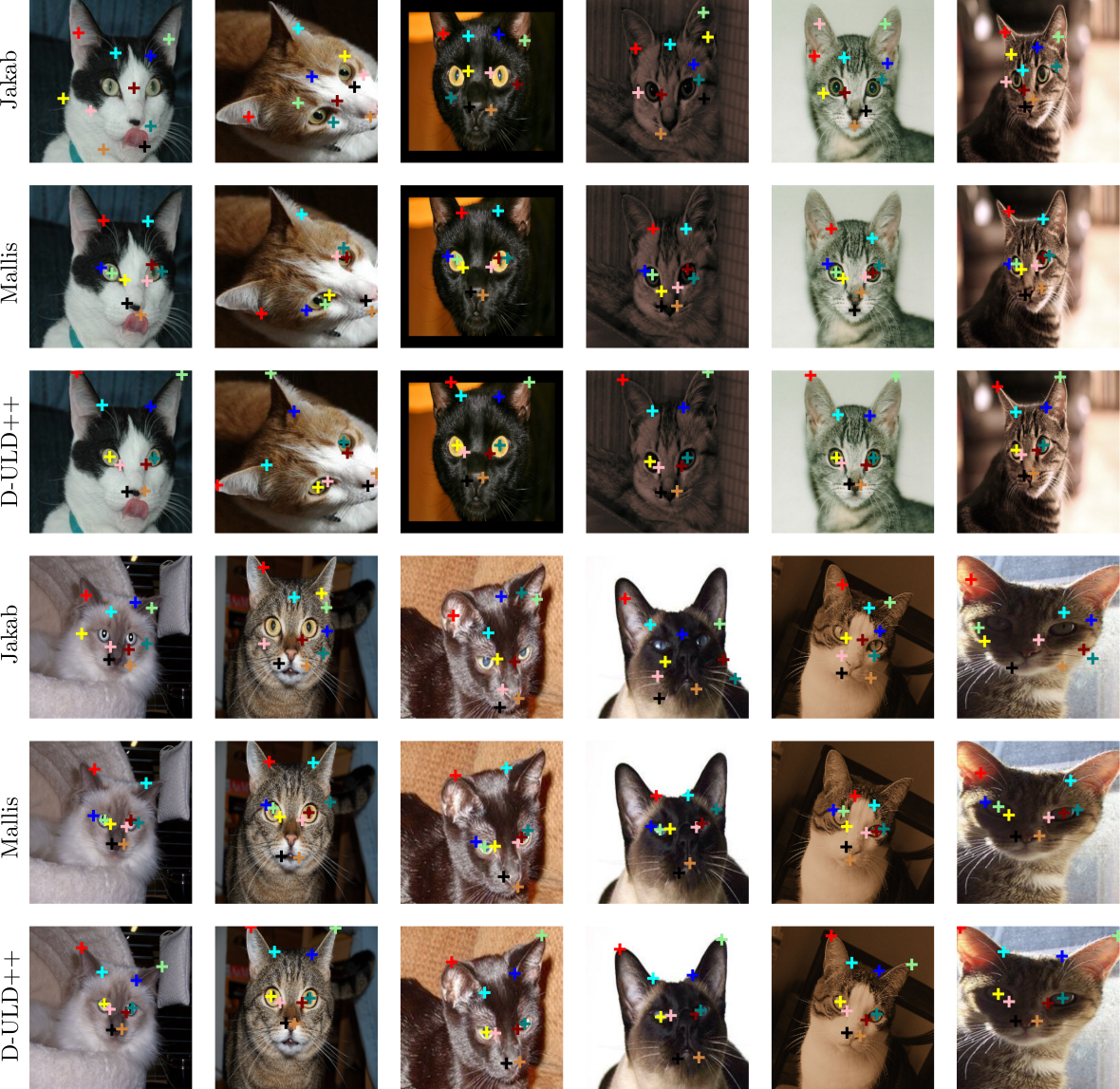}
    \caption{Results Comparison on CatHeads for Jakab, Mallis and D-ULD++.}
    \label{fig:catheads-results-supp}
\end{figure*}

\begin{figure*}
    \centering
    \includegraphics[scale=0.9]{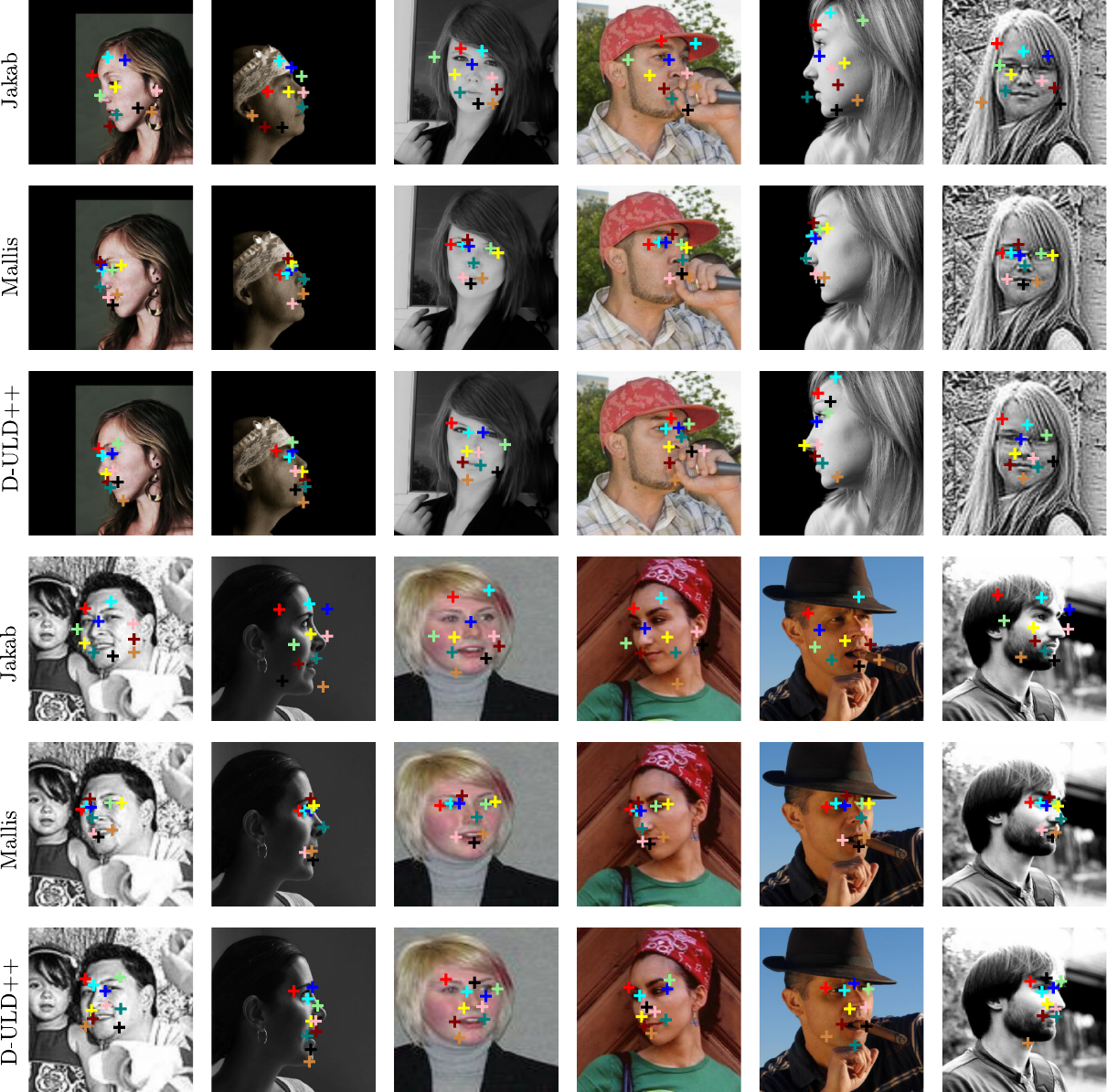}
    \caption{Results Comparison on LS3D for Jakab, Mallis and D-ULD++.}
    \label{fig:ls3d-results-supp}
\end{figure*}

We show additional qualitative results for LS3D~(\Cref{fig:ls3d-results-supp}), CatHeads~(\Cref{fig:catheads-results-supp}) and AFLW~(\Cref{fig:aflw-results-supp}) comparing 3 methods, Jakab, Mallis and D-ULD++. Jakab~\cite{jakab2018unsupervised} generally learns landmarks with poor localization, occasionally not even lying in the image ROI. Mallis~\cite{mallis2023keypoints} performs much better localizing most landmarks well, but a few landmarks are still in smooth regions that lack distinctive edges and are thus poorly localized. Finally, D-ULD++ is reliably able to localize landmarks that are lying in image regions with distinctive edges. 

\end{document}